\documentclass[nohyperref]{article}

\usepackage{microtype}
\usepackage{graphicx}
\usepackage{subfigure}
\usepackage{booktabs} 

\usepackage{hyperref}

\usepackage{url}            
\usepackage{booktabs}       
\usepackage{amsfonts}       
\usepackage{nicefrac}       
\usepackage{microtype}      
\usepackage{xcolor}         
\usepackage{bm}
\usepackage{dsfont} 
\usepackage{wrapfig}
\usepackage{cancel}
\usepackage{graphicx}

\usepackage[accepted]{icml2022}

\usepackage{amsmath}
\usepackage{amssymb}
\usepackage{mathtools}
\usepackage{amsthm}

\usepackage[capitalize,noabbrev]{cleveref}

\theoremstyle{plain}

\theoremstyle{definition}

\theoremstyle{remark}

\usepackage[textsize=tiny]{todonotes}

\icmltitlerunning{Function-space Inference with Sparse Implicit Processes}

\begin{document}

\twocolumn[
\icmltitle{Function-space Inference with Sparse Implicit Processes}




\begin{icmlauthorlist}
\icmlauthor{Simón Rodríguez Santana}{icmat}
\icmlauthor{Bryan Zaldivar}{uv}
\icmlauthor{Daniel Hernández-Lobato}{uam}
\end{icmlauthorlist}

\icmlaffiliation{icmat}{Institute of Mathematical Sciences (ICMAT-CSIC), Madrid, Spain.}
\icmlaffiliation{uam}{Escuela Polit\'ecnica Superior, Universidad Aut\'onoma de Madrid, Spain.}
\icmlaffiliation{uv}{Institute of Corpuscular Physics, University of Valencia and CSIC, Spain.}

\icmlcorrespondingauthor{Simón Rodríguez Santana}{simon.rodriguez@icmat.es}

\icmlkeywords{Machine Learning, ICML, Bayesian Machine Learning, Approximate Inference, Stochastic Processes, Implicit Processes, Model Bias}

\vskip 0.3in
]



\printAffiliationsAndNotice{}  

\begin{abstract}
Implicit Processes (IPs) represent a flexible framework that can be used to describe a wide variety of models, from Bayesian neural networks, neural samplers and data generators to many others. IPs also allow for approximate inference in function-space. This change of formulation solves intrinsic degenerate problems of parameter-space approximate inference concerning the high number of parameters and their strong dependencies in large models. For this, previous works in the literature have attempted to employ IPs both to set up the prior and to approximate the resulting posterior. However, this has proven to be a challenging task. Existing methods that can tune the prior IP result in a Gaussian predictive distribution, which fails to capture important data patterns. By contrast, methods producing flexible predictive distributions by using another IP to approximate the posterior process cannot tune the prior IP to the observed data. We propose here the first method that can accomplish both goals. For this, we rely on an inducing-point representation of the prior IP, as often done in the context of sparse Gaussian processes. The result is a scalable method for approximate inference with IPs that can tune the prior IP parameters to the data, and that provides accurate non-Gaussian predictive distributions.
\end{abstract}


\section{Introduction}
\label{sec:introduction}

Approximate inference has recently attracted an enormous interest from the 
neural networks (NNs) community due to the high performance and 
popularity of these models \cite{gal2016dropout, sun2019functional}. Specifically, approximate inference is seen as a fast and approachable way for obtaining more complete information about the confidence on the predictions made by NNs \cite{blundell2015weight, gal2016uncertainty}. The resulting methods are more informative, which can be useful in real-world scenarios. This is specially important in sensitive tasks that may require taking decisions based on the obtained predictions and their respective uncertainties, \emph{e.g.}, improving the safety of autonomous vehicles \cite{mcallister2017concrete}. Employing the Bayesian framework, in which one uses probability distributions to assign degrees of belief on the weight values of these models, can be successful in this task \cite{BlundellCKW15, graves2011practical}. However, most of these techniques come with intrinsic difficulties such as the choice of a meaningful prior distribution or the intractability of some of the expressions needed, which must be approximated. Moreover, the increasing complexity of models can also lead to difficulties, since performing inference in the space of parameters becomes more challenging in larger models. In such a case, the parameter space presents numerous symmetries and strong dependencies, which makes approximating the target parameter posterior distribution a complicated task. This makes the models harder to train, requiring more data to do so, and compromises 
the prediction performance \cite{sun2019functional}. 

There has been an increasing interest 
recently in \textit{implicit stochastic processes} (IPs) as a way to express prior 
distributions on the function space to face the problems described \cite{ma2019variational,sun2019functional}. Although it 
requires a slightly more complex formulation, doing approximate inference in the function space 
presents a way to simplify the problems that arise in the aforementioned parameter-space. Nevertheless, recent work has only been able to show partial success in this matter: the proposed methods so far have only been able to update the prior model parameters according to the data by sacrificing the flexibility of the predictive distribution, which is fixed to be Gaussian. Alternatively, they can only generate a more complex and flexible predictive distribution at the expense of not being capable of adjusting the prior parameters. Thus, a general method that can do both things is missing in the literature. Namely, 
(i) Update the prior to learn important features of the data.
(ii) Provide non-Gaussian and flexible predictive distributions.

Our contribution is a new method for approximate inference using IPs called \emph{Sparse Implicit Process} (SIP). SIP fulfills both objectives (i) and (ii). We have evaluated SIP in the context of Bayesian NNs and neural samplers \cite{ma2019variational}, as two illustrative examples to showcase its flexibility. Importantly, however, IPs and our method, SIP, are very general models and include, as particular cases, other commonly used priors over functions. Some examples include \textit{normalizing flows} \cite{rezende2015variational}, deep GPs \cite{damianou2013deep} and warped GPs \cite{snelson2004warped}. To obtain a scalable method that can address very large datasets we consider an \textit{inducing-point} approach in SIP in which the number of latent variables on which to perform inference is reduced from $N$ to $M\ll N$, with $N$ the training set size \cite{snelson2005sparse, titsias2009variational}. We have evaluated SIP in extensive regression experiments, including synthetic and real-world problems. When compared against other methods from the literature, SIP often leads to better generalization properties and it captures complex patterns in the predictive distribution. It can also adjust the prior parameters to explain the training data. Finally, in very large datasets, it also remains scalable, reaching good results in less time than other methods.

\section{Background}
\label{sec:background}

We describe first current methods for approximate inference in parameter-space using implicit distributions.
After that, we explain in detail how to carry out approximate inference in the functional space using IPs.

\subsection{Parameter-space Approximate Inference using Implicit Distributions}
\label{subsec:weight_space_optimization_with_implicit_distributions}

Consider some data $\mathcal{D}=\{\mathbf{x}_i,y_i\}_{i=1}^N$, a
 prior $p(\mathbf{w})$ over the model parameters $\mathbf{w}$ and
a likelihood $p(\mathbf{y}|\mathbf{w},\mathbf{X})$. The goal is to
find a distribution $q_\phi(\mathbf{w})$ to approximate the exact
posterior $p(\mathbf{w}|\mathbf{y},\mathbf{X})$. In variational inference, 
$q$ maximizes the evidence lower bound (ELBO) \cite{jordan1999introduction}:
\begin{align}
\mathcal{L}(\phi) &= 
\mathds{E}_{q_\phi(\mathbf{w})}[\log p(\mathbf{y}|\mathbf{w},\mathbf{X})] - \text{KL}(q_\phi(\mathbf{w})|p(\mathbf{w}))
        \label{eq:lower_bound_vi}
\end{align}
where $\text{KL}(\cdot, \cdot)$ is the Kullback-Leibler divergence between distributions.
Maximizing (\ref{eq:lower_bound_vi}) is equivalent to minimizing 
$\text{KL}(q_\phi(\mathbf{w})|p(\mathbf{w}|\mathbf{y},\mathbf{X}))$.
Typical approaches for Bayesian NN (BNN) models use a parametric distribution $q$
that assumes independence among the components of $\mathbf{w}$ \cite{BlundellCKW15, graves2011practical}.
By contrast, in \citet{mescheder2017adversarial,santana2020adversarial}, a more expressive 
approximation is used, making $q$ implicit, \emph{i.e.}, $q_\phi(\mathbf{w}) = \int q_\phi(\mathbf{w}|\bm{\epsilon})
p(\bm{\epsilon}) d \bm{\epsilon}$, with $\bm{\epsilon}$ some random noise. 
Therefore, if $\bm{\epsilon}$ is high-dimensional and $q_\phi(\mathbf{w}|\bm{\epsilon})$
is complicated enough, the result is a very flexible approximate distribution $q_\phi(\mathbf{w})$.
Note, however, that although the first term in (\ref{eq:lower_bound_vi}) can be estimated via Monte Carlo sampling,
the KL contribution is intractable since $q$ lacks a closed-form density.
To solve this, the KL term can be re-written as:
\begin{align}
        \text{KL}(q_\phi(\mathbf{w})|p(\mathbf{w})) & =
        \mathds{E}_{q_\phi(\mathbf{w})}\left[ \log q_\phi(\mathbf{w}) - \log p(\mathbf{w}) \right]\nonumber \\
	&  = \mathds{E}_{q_\phi(\mathbf{w})}\left[ T(\mathbf{w}) \right]\,,
        \label{eq:KL_rewriting}
\end{align}
where $ T(\mathbf{w})$ is the log-ratio between $q_\phi(\mathbf{w})$ and the prior for $\mathbf{w}$. 
Let $\sigma(\cdot )$ be the sigmoid function. $T(\mathbf{w})$ can be estimated using a classifier 
$T_\omega(\mathbf{w})$, with parameters $\omega$ obtained by maximizing the objective:
\begin{align}
\label{eq:objective_discriminator_general}
    \mathds{E}_{q_\phi} \left[\log \sigma (T_\omega(\mathbf{w}))\right] + \mathds{E}_{p(\mathbf{w})}  \left[ \log (1 
	- \sigma (T_\omega(\mathbf{w}))) \right]\,.
\end{align}
If the discriminator is flexible enough, its optimal value, $T_{\omega^\star}$, is exactly the log ratio between 
$q_\phi(\mathbf{w})$ and $p(\mathbf{w})$, \emph{i.e.}, $ T_{\omega^\star} (\mathbf{w}) = \log q_\phi(\mathbf{w}) - \log p(\mathbf{w})$ \cite{mescheder2017adversarial}. Using this, (\ref{eq:lower_bound_vi}) becomes
\begin{align}
\mathcal{L}(\phi) & = \sum_{i=1}^N \mathds{E}_{q_\phi}[\log p(y_i|\mathbf{w},\mathbf{x}_i)]  
	- \mathds{E}_{q_\phi}[T_{\omega^\star}(\mathbf{w})] \,,
        \label{eq:lower_bound_avb}
\end{align}
where the KL term is replaced by the discriminator, which is trained simultaneously.
This enables the use of an implicit model for $q_\phi(\mathbf{w})$. 
Finally, instead of minimizing the regular KL-divergence between $q$ and 
the posterior, other approaches suggest to minimize a more general $\alpha$-divergence, which
includes the KL-divergence as a particular case \cite{hernandez2016black}. This has 
shown to overall improve the final results \cite{santana2020adversarial, ma2019variational}. 

\subsection{Function-Space Approximate Inference}
\label{sec:func_space_approx_inf}

Implicit processes (IPs) are defined as a collection of random
variables $f(\cdot)$ such that any finite set of evaluations
$(f(\mathbf{x}_1), \cdots , f(\mathbf{x}_N))^\text{T}$ has joint distribution determined by
the generative process \cite{ma2019variational}:
\begin{align}
\mathbf{z} \sim p(\mathbf{z})\,,& \quad  f(\mathbf{x}_i) = g_\theta (\mathbf{x}_i, \mathbf{z}), \quad
        \forall \mathbf{x}_i \in \mathcal{X},
\end{align}
where $\mathbf{z}$ is some random variable that summarizes the randomness,
$\theta$ represents the parameters of the process and $\mathcal{X}$ is the input space.
We use the notation $f(\cdot) \sim \mathcal{IP}(g_\theta(\cdot, \cdot), p_{\mathbf{z}})$ to indicate that $f$ is sampled from the corresponding IP with parameters $\theta$, using samples from $p(\mathbf{z})$ (denoted as $p_{\mathbf{z}}$).
This definition of IPs results in a framework that is general enough to include many different models.
For example, Bayesian NNs can be described using IPs if the randomness is given by the prior $p(\mathbf{w})$ over 
the NN weights $\mathbf{w}$.  We would then sample a function parameterized by $\mathbf{w} \sim p(\mathbf{w})$, which specifies
the output of the NN as $f(\mathbf{x}) = g_\theta (\mathbf{x}, \mathbf{w})$ for every 
$\mathbf{x} \in \mathcal{X}$. $\theta$ are here the parameters of $p(\mathbf{w})$. If $p(\mathbf{w})$ is a factorizing
Gaussian, $\theta$ will be the corresponding means and variances.
Other important models that can be described as IPs include, \emph{e.g.},
neural samplers (NS), warped GPs, and deep GPs \cite{ma2019variational,snelson2004warped,damianou2013deep}.
Previous works using IPs for inference are the variational
implicit processes (VIP) and the functional variational Bayesian NN (fBNN)
\cite{ma2019variational, sun2019functional}. We introduce them next.

VIPs approximate the marginal likelihood of the prior IP by the marginal likelihood of a GP.
For this,  an empirical covariance function is estimated by sampling from the prior IP. 
Namely, $f_s^\theta(\cdot) \sim \mathcal{IP}(g_\theta(\cdot, \cdot), p_{\mathbf{z}})$. Then,
the prior mean and covariances of the GP are:
\begin{equation}
\begin{aligned}
\label{eq:moments_GP_solution_VIP}
        m^\star(\mathbf{x}) & =  \frac{1}{S} \sum_{s=1}^S f_s^\theta(\mathbf{x})\,, \\
        \mathcal{K}^\star(\mathbf{x}_1,\mathbf{x}_2) & =  \frac{1}{S} \sum_{s=1}^S
        \Delta_s^\theta(\mathbf{x}_1) \Delta_s^\theta(\mathbf{x}_2)\,,
\end{aligned}
\end{equation}
where $\Delta_s(\mathbf{x}) = f_s^\theta(\mathbf{x}) - m^\star(\mathbf{x})$.
They then approximate  $p(\mathbf{y}|\mathbf{X},\theta)$ by $q_\text{GP}(\mathbf{y}|\mathbf{X},\theta)$,
\emph{i.e.}, the marginal likelihood of a GP with the estimated means and covariances,
and maximize the latter, expecting that it will increase $p(\mathbf{y}|\mathbf{X},\theta)$ as well.
To guarantee scalability and avoid the cubic cost of the GP they further approximate the GP using
a linear model with the same mean and covariances. The linear model is efficiently tuned by optimizing
the $\alpha$-energy function, performing gradient descent w.r.t. $\theta$ via the method described in
\citet{hernandez2016black}. A limitation of VIP, however, is that the final predictive distribution is
Gaussian (that of a GP) which may lack enough flexibility.

By contrast, fBNNs rely on VI and, instead of using a GP, they use another IP to approximate 
the posterior of the prior IP \cite{sun2019functional}. In the ELBO in 
(\ref{eq:lower_bound_vi}), the first term can be estimated by
Monte Carlo when using an IP as the approximate posterior.  However, the KL term becomes
the KL-divergence between stochastic process, which is intractable.
To address this, fBNN evaluates the KL-divergence between distributions at
a finite set of points $\tilde{\mathbf{X}}$ drawn at random from the input space, 
leaving the ELBO as:
\begin{align}
\label{eq:fbnn_elbo}
                \mathcal{L}(q) & =  
        \mathds{E}_q[\log p(\mathbf{y}|\mathbf{f}^\mathbf{X})] - 
	\mathds{E}_{\tilde{\mathbf{X}}}[\text{KL}(q(\mathbf{f}^{\tilde{\mathbf{X}}})|p(\mathbf{f}^{\tilde{\mathbf{X}}}))]\,,
\end{align}
where $\mathbf{f}^\mathbf{X}$ and $\mathbf{f}^{\tilde{\mathbf{X}}}$
are the IP values at 
$\mathbf{X}$ and $\tilde{\mathbf{X}}$, respectively. This objective function is then maximized w.r.t 
the parameters of the posterior approximate IP $q$.  Critically, $\tilde{\mathbf{X}}$ must 
cover training and testing regions of the input space. Therefore, fBNN may suffer in large
dimensional datasets. Moreover, fBNN is
unable to fit the prior IP model by itself due to estimating
the gradients of the KL-divergence term using a spectral gradient estimator \cite{sun2019functional}.

\section{Sparse Implicit Processes}

We introduce \textit{Sparse Implicit Processes} (SIP), a new method for approximate inference when using IPs.
As in fBNN, in SIP we consider another IP to approximate the posterior process resulting from combining the 
prior IP with the data using Bayes' rule.
However, unlike in fBNN we perform inference on finite sets of variables, as it is usually
done with GPs. This avoids the problem of computing the KL-divergence between stochastic processes.
SIP also allows to easily adjust the parameters of the prior IP.
However, we will need to address two issues: (i) avoid the number of latent variables
increasing with the number of training points $N$; and (ii) deal with the intractability
of the computations.

The first problem can be sorted out by considering an approximation based on inducing 
points, as in sparse GPs \cite{snelson2005sparse,titsias2009variational}.
Instead of making inference about $\mathbf{f}=(f(\mathbf{x}_1,),\ldots,f(\mathbf{x}_N))^\text{T}$
we perform inference about the process values at $M \ll N$ inducing points.
We denote the set of inducing points $\overline{\mathbf{X}}$, and denote the IP values at these
input locations $\mathbf{u}=(f(\overline{\mathbf{x}}_1),\ldots,f(\overline{\mathbf{x}}_M))^\text{T}$.
Next, we focus on approximating $p(\mathbf{f},\mathbf{u}|\mathcal{D})$, which only depends on finite sets 
of variables. For this, we consider the approximate posterior distribution:
\begin{align}
\label{eq:posterior_approximation}
    q(\mathbf{f}, \mathbf{u}) &= p_\theta(\mathbf{f}| \mathbf{u}) q_\phi(\mathbf{u})\,,
\end{align}
where $q_\phi(\mathbf{u})$ is an implicit distribution with parameters $\phi$, and $\theta$ are the
prior IP parameters. Critically, $p_\theta(\mathbf{f}| \mathbf{u})$ is fixed and 
$q_\phi(\mathbf{u})$ is tunable, as in the variational sparse GP approximation \cite{titsias2009variational}. We will use a Monte Carlo GP approximation of $p_\theta(\mathbf{f}| \mathbf{u})$, following what is done in VIP to approximate the prior IP using a GP. However, unlike VIP, here we only use the Gaussian approximation in part of the posterior. We will explicitly define $p_\theta(\mathbf{f}| \mathbf{u})$ later. For now, using this decomposition, the obtained functional-ELBO is:
\begin{align}
\label{eq:fELBO_SIP}
        \mathcal{L}(\phi,\theta) & = 
	\mathds{E}_{q_{\phi,\theta}} \left[ \log \frac{p(\mathbf{y}|\mathbf{f})
	\cancel{p_\theta(\mathbf{f}|\mathbf{u})}p_\theta(\mathbf{u})}{\cancel{p_\theta(\mathbf{f}|\mathbf{u})}q_\phi(\mathbf{u})} \right]
	\nonumber \\
	& =  \mathds{E}_{q_{\phi}}[ \log p(\mathbf{y}|\mathbf{f}) ] - \text{KL}(q_\phi(\mathbf{u})|p_\theta(\mathbf{u}))\,.
\end{align}
The first term can be estimated by Monte Carlo sampling, as in standard approaches for VI. However,
the second term lacks any closed-form solution, since it is the KL-divergence between two implicit 
distributions. To estimate it we rely on the method described
in Section \ref{subsec:weight_space_optimization_with_implicit_distributions}, where a classifier is used
to estimate the log-ratio \cite{mescheder2017adversarial, santana2020adversarial}.
Namely,
\begin{align}
        \text{KL}(q_\phi(\mathbf{u})|p_\theta(\mathbf{u})) & = 
        \mathds{E}_q \left[ T_{\omega^\star}(\mathbf{u}) \right]\,,
\end{align}
where $T_{\omega^\star}(\mathbf{u})$ is approximated by a NN 
that discriminates between samples from $q_\phi(\mathbf{u})$ and $p_\theta(\mathbf{u})$.
Note, however, that $T_{\omega^\star}(\mathbf{u})$ depends on $\phi$ and $\theta$. Nevertheless, as
argued in \citet{mescheder2017adversarial}, $\mathds{E}_q (\nabla_\phi [T_{\omega^\star}(\mathbf{u})])=0$
if $T_{\omega^\star}(\mathbf{u})$ is optimal. During the training procedure, we must ensure that the discriminator is trained enough so that it provides a fitting approximation to the original KL term in \eqref{eq:fELBO_SIP} (see Appendix \ref{app:adaptive_contrast} for further discussion on the matter).

Regarding $\nabla_\theta T_{\omega^\star}(\mathbf{u})$, note that it is 
expected to be small when compared to $\nabla_\theta \mathds{E}_{q_{\phi}}[ \log p(\mathbf{y}|\mathbf{f}) ]$ (see Appendix \ref{app:gradient_analysis}). However, although small, these gradients play a crucial role to
adjust the prior distribution to the observed data. This can be critical in different cases, 
\emph{e.g.} when specifying a prior is complicated, or when making predictions far from the 
observed data. When a classifier is used to approximate the KL-divergence, these gradients will be
ignored, which results in not properly adjusting the prior parameters $\theta$ (see Appendix 
\ref{app:gradient_analysis}). To partially correct for this we substitute 
the KL-divergence by the symmetrized KL-divergence: 
\begin{align}
        \text{KL}(q_\phi|p_\theta) & \approx
	\frac{1}{2}(\text{KL}(q_\phi|p_\theta) +\text{KL}(p_\theta|q_\phi))
	\,,
	\label{eq:kl_approx}
\end{align}
where we have omitted the dependence on $\mathbf{u}$ of $q_\phi$ and $p_\theta$.
The KL-divergence in VI is just a regularizer enforcing $q$ to look similar to the prior. 
(\ref{eq:kl_approx}) is expected to play a similar role.
Moreover, modifying this regularizer in VI is a common
approach that often gives better results. See, \emph{e.g.}, \citet{wenzel20a}.
Importantly, the reversed KL-divergence, $\text{KL}(p_\theta|q_\phi)$, involves
also the log-ratio between the prior and the posterior approximation $q$. Therefore,
it can also be estimated using the same classifier $T_{\omega^\star}(\mathbf{u})$.
Critically, however, it involves an expectation with respect to $p_\theta(\mathbf{u})$.
This introduces some easy to compute gradients with respect to the parameters of the prior $\theta$.
As shown in our experiments, we have found those dependencies to be enough to provide some prior adaptation to
the data. This choice is also supported by good empirical results obtained and because 
the prior adaptation is not observed when only the first KL-divergence term is considered (see Appendix \ref{app:gradient_analysis}). 
When changing the KL-term the objective is no longer a lower bound on the log-marginal likelihood.
However, this is also the case for VIP, which uses a GP approximation to the log-marginal likelihood \cite{ma2019variational}. Moreover, as indicated by \citet{knoblauch2019generalized}, the distributions obtained, as alternative approximations to VI, often perform better in practice. Note that VI is optimal only relative to a particular objective, whose origins are assumptions (\emph{e.g.}, formulation of the likelihood and the prior) that could be misaligned with reality. 

An important remark is that the data-dependent term in (\ref{eq:fELBO_SIP})
only considers the squared error (in the case of a Gaussian likelihood).
We follow \citet{santana2020adversarial} and resort to the
energy function employed in BB-$\alpha$ and VIP 
\cite{hernandez2016black,ma2019variational}. Replacing the data-dependent
term in (\ref{eq:fELBO_SIP}) results in the approximate optimization of
$\alpha$-divergences. This generalizes the original data-dependent term, making the objective more sensitive to the actual distribution of the data for higher values of $\alpha$. In particular, for
$\alpha \rightarrow 0$, the data-dependent term in \eqref{eq:fELBO_SIP} is obtained, while for $\alpha \rightarrow 1$, it resembles the log-likelihood of the training data. With this re-formulation, the objective depends on $\alpha$ and is:
\begin{align}
\label{eq:alpha_f-elbo}
    \mathcal{L}_\alpha^\star(\phi,\theta) & = \frac{1}{\alpha} \sum_{i=1}^N \log \mathds{E}_{q_{\phi,\theta}} 
	[p(y_i|f_i)^\alpha ]  \nonumber \\
	& \quad - \frac{1}{2} \left[ \text{KL}(q_\phi | p_\theta) 
	+ \text{KL}(p_\theta | q_\phi) \right]\,,
\end{align}
where the first term can be estimated via Monte Carlo, using a small number of samples as 
in \citet{hernandez2016black}. Moreover, $\alpha$ can be chosen to target the data log-likelihood, \emph{i.e.},
when $\alpha=1$. When $\alpha \rightarrow 0$ the original data-dependent term in (\ref{eq:fELBO_SIP}) is obtained.
The bias introduced by the $\log(\cdot)$ function in (\ref{eq:alpha_f-elbo}) becomes 
negligible by using a small number of samples \cite{hernandez2016black}.

The other critical point of SIP is how to compute
$p_\theta(\mathbf{f}|\mathbf{u})$ in \eqref{eq:posterior_approximation}. We approximate this conditional
distribution by the conditional of a GP with the same covariance and mean function as the prior IP, as it is done in VIP
\cite{ma2019variational}. More precisely, given samples of $\mathbf{f}$ and $\mathbf{u}$, we employ
(\ref{eq:moments_GP_solution_VIP}) to estimate the means and covariances needed via Monte Carlo.
We then resort to the GP predictive equations described in \citet{rasmussen2005book}. Namely,
$p_\theta(\mathbf{f}|\mathbf{u})$ is approximated as Gaussian with mean and covariance
\begin{align}
        \mathds{E}[\mathbf{f}] & = m(\mathbf{x}) +
        \mathbf{K}_{\mathbf{f},\mathbf{u}}(\mathbf{K}_{\mathbf{u},\mathbf{u}} + \mathbf{I} \sigma^2)^{-1}
        (\mathbf{u}-m(\overline{\mathbf{X}}))\,,
        \nonumber \\
        \text{Cov}(\mathbf{f}) &= \mathbf{K}_{\mathbf{f},\mathbf{f}} -
        \mathbf{K}_{\mathbf{f},\mathbf{u}}(\mathbf{K}_{\mathbf{u},\mathbf{u}} + \mathbf{I} \sigma^2)^{-1} \mathbf{K}_{\mathbf{u},\mathbf{f}},
\end{align}
with $\sigma^2$ a small noise variance (set to $10^{-5}$). Moreover, 
$m(\cdot)$ and each entry in $\mathbf{K}_{\mathbf{f},\mathbf{u}}$ and
$\mathbf{K}_{\mathbf{u},\mathbf{u}}$ is estimated empirically using (\ref{eq:moments_GP_solution_VIP}), 
as in VIP \cite{ma2019variational}.

Finally, predictions at a new point $\mathbf{x}^\star$ are estimated via Monte Carlo with $S$ samples.
Let $\mathbf{u}_s \sim q_\phi(\mathbf{u})$. Then,
\begin{align}
\label{eq:predictions_SIP}
        p(f(\mathbf{x}_\star)|\mathbf{y},\mathbf{X}) \approx  \frac{1}{S} \sum_{s=1}^S p_\theta(f(\mathbf{x}_\star)|
        \mathbf{u}_s)\,.
\end{align}
Thus, the predictive distribution is a mixture of Gaussians, where each Gaussian is 
determined by one sample extracted from the approximate posterior IP at $\overline{\mathbf{X}}$.
This means that SIP can produce flexible predictive distributions that need not
be Gaussian, unlike VIP.

\begin{figure}[th!]
    \centering
    \includegraphics[width=0.47\textwidth]{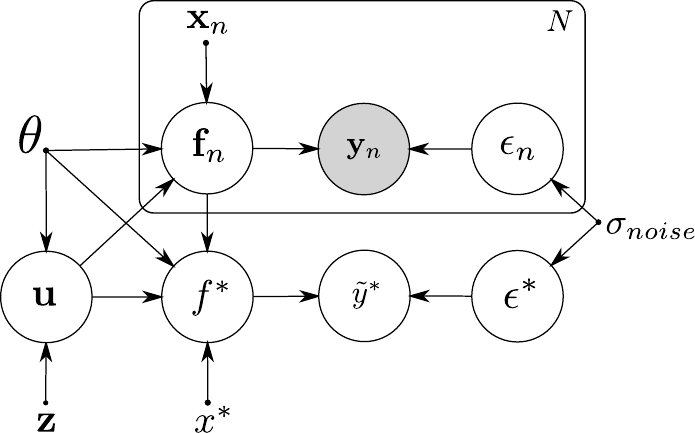}
    \caption{Graph model for the SIP method. Point-like vertices denote deterministic variables and circles represent random variables, which can either be observed (\textit{grey}) or unobserved (\textit{white}).}
    \label{fig:graph_model}
\end{figure}

To conclude, we illustrate the SIP method with its graphical model in Figure \ref{fig:graph_model}. 
Here $\theta$ are the prior IP parameters. The inducing points' locations are 
explicitly depicted as $\mathbf{z}$, with their respective sampled functions values locations 
$\mathbf{u}$. The function values on the input data are $\mathbf{f}_n$, with $N$ training 
data points $\{\mathbf{x}_n, \: n = 1 , \: .\,.\,. \: , N\}$, 
while $f^*$ is used for the test data ($\mathbf{x}^*$). Finally, each noise contribution 
is sampled from $\mathcal{N}(0, \sigma_{noise})$ (\emph{i.e.} $\epsilon_n$ for 
training and $\epsilon^*$ for testing). We then estimate the output, which 
can correspond to an observed ($\mathbf{y}_n$, in training) or unobserved variable 
($\tilde{y}^*$, in testing).



\section{Related Work}
\label{sec:background_and_related_work}

Traditionally, approximate inference has been conducted in parameter space.
Bayesian Neural Networks (BNNs) are a notable example. They enable predictive 
distributions accounting for prediction uncertainty in the context of NN. A
widely explored technique for this is Bayes by back-propagation (BBB)
\cite{BlundellCKW15, graves2011practical, jordan1999introduction}. BBB performs
variational inference (VI) in the space of weights using a parametric distribution
\cite{mackay1992practical, graves2011practical}. Novel approaches are leading to 
new interpretations and generalizations based on VI, from which the resulting methods 
can have appealing theoretical properties \cite{knoblauch2019generalized}. However, constraining 
the approximate solution to a certain parametric family which in most cases assumes independence may be
too restrictive. More precisely, in complex NNs this approach leads to pathological
behaviors that may result in a poor generalization \cite{sun2019functional}. An alternative to BBB
is Probabilistic Back-propagation (PBP) \cite{hernandez2015probabilistic}. This method propagates 
probabilities through the NN. PBP has proven to be efficient and scalable, although it also has 
limited expressiveness in the posterior approximation, which is a factorizing Gaussian. Therefore, 
as BBB, PBP suffers from problematic approximation bias \cite{graves2011practical,BlundellCKW15}.

Recent works have analyzed the properties of simple variational approximation methods. In 
\citet{foong2019expressiveness} they are shown to underestimate the prediction uncertainty. 
Wide BNNs are also subject of study in \citet{coker2021wide} under the mean-field assumption. 
Pathological behaviors arise for deeper BNN models in the approximate posterior, which strongly 
differs from the exact one. Therefore, simple VI approximations based on, \emph{e.g.}, mean-field 
should be avoided.  More flexible approximations can be obtained using normalizing flows (NF) 
\cite{rezende2015variational}. NFs perform a series of non-linear invertible transformations on 
the variables of a tractable parametric distribution, obtaining a more complex distribution with 
closed-form density. For this, the transformations must be invertible, which may limit the flexibility of 
the approximate solution.

There has been a growing interest in increasing the flexibility of the 
approximate distribution \cite{liu2016stein,salimans2015markov,tran2017hierarchical}.
An implicit model can be useful for this 
\cite{li2016wild, mescheder2017adversarial, santana2020adversarial}.
There, the approximate distribution lacks a closed-form density, but
one can easily sample from it. Since $q$ lacks a density expression, it becomes difficult 
to evaluate the KL-divergence term of the VI objective.  Adversarial Variational Bayes (AVB) solves this 
problem \cite{mescheder2017adversarial} using an auxiliary discriminator.
See Section \ref{subsec:weight_space_optimization_with_implicit_distributions} for further details.
AVB has been extended by Adversarial $\alpha$-divergence minimization
(AADM), which combines the BB-$\alpha$ objective \cite{hernandez2016black}
with an implicit model for $q$ to locally minimize an $\alpha$-divergence \cite{santana2020adversarial}.
In AADM $\alpha\in (0,1]$ is an adjustable parameter which allows to interpolate between targeting the direct 
and the reversed KL-divergence. Furthermore, AADM can model complex predictive distributions, unlike AVB.

Other  works consider using inducing points and sparse models in the context of BNN 
\cite{immer2021improving, ritter2021sparse}. However, in contrast to SIP, the locations of 
the inducing points are fixed, and also, unlike SIP, approximate inference is carried 
out in the parameter space.  SIP is also more general since it is not restricted to work with BNNs.

The methods described so far suffer from the problems of working in the space of parameters,
which is high-dimensional and includes strong dependencies. Recent works have shown better results 
by performing approximate inference in the space of functions \cite{ma2019variational,sun2019functional}. 
Characterizing function-space inference, as well as constructing effective frameworks for it 
focuses many research efforts nowadays \cite{burt2020understanding}. In some of the most successful 
approaches, the underlying model is constructed using an implicit stochastic process. 
Two successful methods for approximate inference in the context of IPs are VIP
\cite{ma2019variational} and fBNN \cite{sun2019functional}, described in detail in Section \ref{sec:func_space_approx_inf}.
VIP is limited to having a Gaussian predictive distribution, which may lack flexibility.
By contrast, in fBNN is difficult to infer the parameters of the prior IP. Moreover, fBNN also relies 
on uniformly covering the input space to guarantee that the posterior IP looks similar to the prior IP in 
regions with no data. This is challenging in high dimensions.
SIP does not have these limitations and can produce flexible predictive distributions
and adjust the prior parameters to the data.

Functional Variational Inference (FVI) is a recent method proposed for approximate
inference in the context of IPs \citep{ma2021}. FVI approximately minimizes 
the KL-divergence between stochastic processes. This is done efficiently in 
a two step approach. First, the prior IP is approximated using a flexible model 
called stochastic process generator (SPG). Then, another SPG is used to efficiently 
approximate the posterior of the prior SPG. Both SPGs share key aspects that 
facilitate this task. In any case, FVI suffers from the same limitation as 
fBNN. It cannot adjust the prior parameters to the data, unlike in SIP. 

In the context of BNNs, \citet{rudner2021} approximate the KL-divergence between 
stochastic processes by considering a NN linearized around its mean parameters. 
After using an inducing point approximation, the result is the KL-divergence between 
two Gaussians. This approach is, however, constrained to (i) using BNNs with 
(ii) a parametric distribution over the weights (\emph{e.g.} Gaussian) and a (iii) mean-field approximation, unlike SIP.

Approximate inference in the functional space is not a new concept.
Methods based on GPs have been used extensively \cite{rasmussen2005book}.
In GPs, the posterior can be computed analytically. However, they have a big 
training cost. Inducing points approximations, similar to those of SIP, can be 
combined with stochastic VI in GPs for better scalability 
\cite{hensman2013}. However, these methods are limited to using 
GP priors and only produce Gaussian predictions. IP based methods, 
such as SIP, can use a wider range of priors and can produce 
non-Gaussian predictions.

\section{Experiments}
\label{sec:experiments}

We evaluate the proposed method, SIP, using different NN models
for the prior: a Bayesian NN (BNN) and 
a neural sampler (NS), both with 2 hidden layers of 50 units.
Such NN size is similar to that used in, \emph{e.g.}, \citet{ma2019variational,sun2019functional,santana2020adversarial}.
In the NS, input noise is standard Gaussian with 10 dimensions. 
The posterior implicit model in SIP uses another NS with 100 noise dimensions.
The tensor-flow code for SIP is included in the \href{https://github.com/simonrsantana/sparse-implicit-processes.git}{github repository}. 
Each method is trained until convergence using a mini-batch size of 10. 
For training, in SIP we use 100 samples to estimate (\ref{eq:alpha_f-elbo}) and its gradients. 
In test, 500 samples are used in (\ref{eq:predictions_SIP}).
The exact number of inducing points used is specified in each experiment, although through preliminary experiments we saw little improvements using more than $50 \, - \, 100$ inducing points. Adding more is expected to improve the accuracy of the method, but also raises the computational cost \cite{titsias2009variational}. Finally, 
we use $\alpha = 1$ in SIP in the synthetic experiments (this targets the 
data log-likelihood for visualization of the predictive distribution) 
and $\alpha = 0.5$ in the real-world ones. $\alpha = 0.5$ gives good 
results for the metrics used and is recommended in 
\citet{santana2020adversarial,ma2019variational}. We compare SIP with 
BBB, VIP, fBNN, and AADM (see Section \ref{sec:background_and_related_work}).
The noise variance is chosen by maximizing the estimate 
of the log-marginal likelihood of each method.
We follow \citet{ma2019variational,sun2019functional}
and focus our experiments on similar regression problems. 

\begin{figure*}[tbh]
    \centering
    \includegraphics[width=1\textwidth]{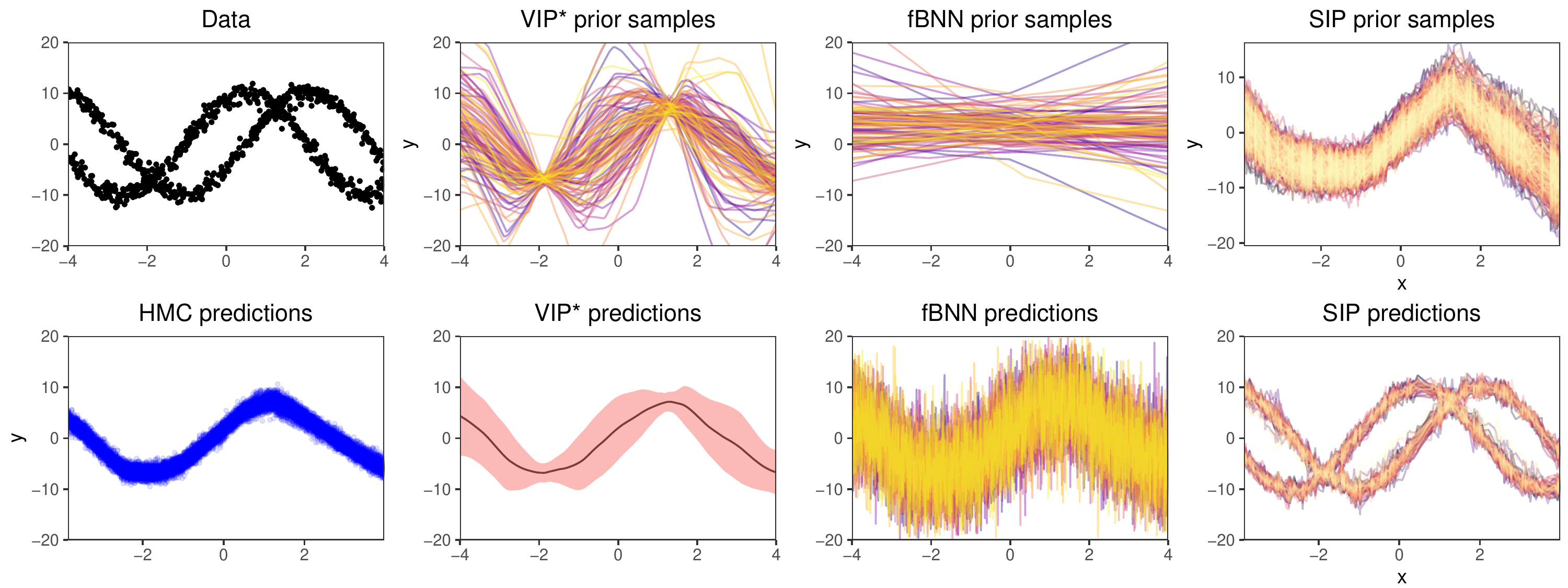}
    \caption{\small Samples from the prior and the predictive distribution of each method. 
	First column contains the data (top row, in black) and the HMC predictions (bottom row, in blue). 
	For the rest of the methods, the first row shows samples from the learned prior distribution. 
	The second row shows the samples from the predictive distribution. Best seen in color.}
    \label{fig:prior_and_predictive_bimodal}
\end{figure*}

\subsection{Synthetic Experiments}
\label{subcsec:synthetic_experiments}

These experiments compare the quality of the predictive distribution given by 
SIP with that of VIP and fBNN.  We train each method using the same BNN prior for 
a fair assessment. Here, in VIP we have disabled the regularization term to favor 
a better prior fit. If included, VIP's heterocedastic behavior is lost (see Appendix \ref{subsec:alternative_bimodal_setups}). 
Moreover, Appendix \ref{app:het_data_experiments} has extra results for another dataset. The results obtained are similar to these 
ones.  Here we also evaluate the performance of Hamilton Monte Carlo (HMC) applied on the same model. The 
prior for HMC is the BNN prior learned by $\text{SIP}$.  The number of inducing points in SIP is set to $50$. 

We first study a bimodal dataset we generated taking 1000 samples of $x$ from a uniform distribution 
$\mathcal{U}(-4, 4)$. Then, we generate  one of two possible values for $y$: 
\begin{align}
    y_1 = 10 \cos(x - 0.5) + \epsilon, \:\:\:\:\:\:\:\:
    y_2 = 10 \sin(x - 0.5) + \epsilon
    \nonumber
\end{align}
with $\epsilon \sim \mathcal{N}(0,1)$. We select either $y_1$ or $y_2$ randomly, 
producing bimodal data. For comparison, we plot the samples from the learned prior 
and the predictive distribution of each method after $2000$ training epochs.

Figure \ref{fig:prior_and_predictive_bimodal} show samples from the learned prior distribution and 
the predictive distribution for $y$ of each method. The original training data is shown in the top left corner.
We observe that in the case of VIP and SIP, the prior model captures the mean value of the training data, 
while  for fBNN seems to have not learned any pattern at all. However, VIP's predictive distribution, which is 
Gaussian, is unable to represent the bimodality of the data. fBNN's predictive distribution, although more flexible 
than the one of VIP by construction, cannot capture the bimodality either. The reason behind this is that the 
data-dependent term of fBNNs focuses on minimizing the squared error, and hence it 
simply outputs the average prediction between the two modes, 
as illustrated in \citet{santana2020adversarial}. In summary, SIP is the only method that learns a 
sensible prior distribution and whose predictions capture the bimodality of the data (see Appendix \ref{app:extra_synth_examples} for extra experiments).

An unexpected result in Figure \ref{fig:prior_and_predictive_bimodal} is that HMC, 
which is expected to be the most accurate method for approximate inference,
cannot capture the bimodal predictions. This is simply because the assumed 
model (a BNN) is wrong and the exact posterior need not be optimal. 
In particular, if one randomly generates functions from the NN prior to 
then contaminate them with additive Gaussian noise, the bimodal predictive distribution is never 
observed, in practice. A uni-modal predictive distribution is obtained. This is the 
one captured by HMC. This makes sense since a BNN prior will converge to a GP when number of hidden units 
is large \citep{neal1994}. A GP will always output a Gaussian predictive distribution and not 
a bi-modal predictive distribution. By contrast, SIP is more flexible and thanks to 
the approximate inference mechanism is able to bypass the wrong model specification 
and produces a more accurate predictive distribution. Therefore, the approximate inference 
method used by SIP is robust to model mis-specification, as a consequence of the data-dependent 
term, which enforces to produce accurate predictions for the training instances.

\begin{table*}[tbh]
\caption{\small Ranking analysis between methods across every multivariate regression problem (lower is better). The 
method with the lowest average rank across data sets and splits is highlighted in bold.}
\begin{center}
\begin{tabular}{l|lll|ll|ll}
Method & BBB & AVB & AADM & $\text{fBNN}_{\text{BNN}}$ & VIP & $\text{SIP}_\text{NS}$ & $\text{SIP}_\text{BNN}$\\
\hline
RMSE & 6.15$ \pm $0.04 & 4.14$ \pm $0.08 & 3.97$ \pm $0.09 & 3.82$ \pm $0.10 & 3.29$ \pm $0.09 & $\mathbf{3.10}$$ \pm $$\mathbf{0.09}$ & 3.53$ \pm $0.08\\
NLL & 5.39$ \pm $0.05 & 3.91$ \pm $0.06 & 2.82$ \pm $0.06 & 5.61$ \pm $0.07 & 3.70$ \pm $0.12 & 3.81$ \pm $0.07 & $\mathbf{2.76}$$ \pm $$\mathbf{0.08}$\\
CRPS & 5.84$ \pm $0.05 & 4.47$ \pm $0.08 & 4.03$ \pm $0.08 & 3.97$ \pm $0.07 & 2.57$ \pm $0.08 & 4.67$ \pm $0.08 & $\mathbf{2.45}$$ \pm $$\mathbf{0.08}$\\
\end{tabular}
\end{center}
\label{tab:ranking_analysis}
 \vspace{-0.15cm}
\end{table*}

\subsubsection{Location of the Inducing Points}
\label{subsec:locations_of_the_inducing_points}

Using inducing points is critical in SIP, and hence it is important to assess that it is able of 
learning their locations, $\overline{\mathbf{X}}$, correctly. 
These points specify the potential values that the posterior IP can take in different regions 
of the input space. More precisely, consider some level of smoothness implied by the prior IP. 
Then, in regions of the input space close to $\overline{\mathbf{X}}$ one should expect a similar process value 
as the one specified by the implicit distribution $q_\phi(\mathbf{u})$. To test the ability to learn the location of 
the inducing points we consider the synthetic dataset in \cite{snelson2005sparse} and 15 inducing points in SIP. We 
initialize them adversarially concentrated in one input location and train the model. In this case we 
do not fit the parameters of $q_\phi$ so that the model focuses on the inducing points locations and the prior 
for prediction, preventing $q_\phi(\mathbf{u})$ to compensate for locations with not enough inducing points nearby.

\begin{figure}[hbt]
    \centering
    \includegraphics[width = 0.47\textwidth]{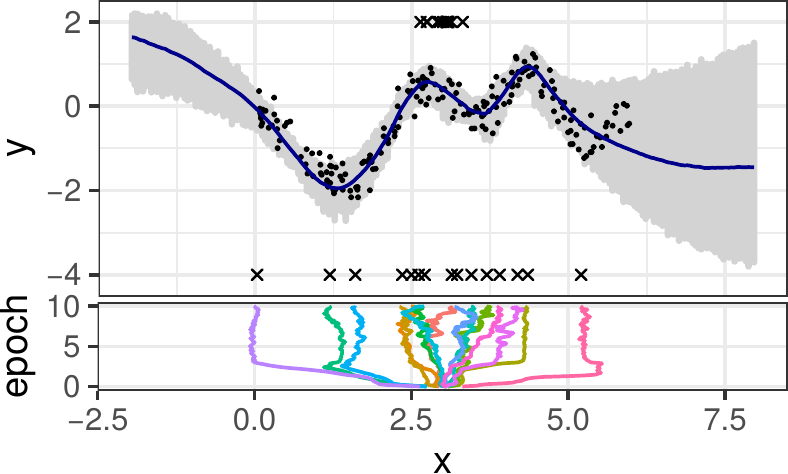}
    \caption{\small Predictive distribution (top) and evolution of the location of the inducing 
	points (bottom) for the dataset in \cite{snelson2005sparse}.  The crosses at the top 
	and bottom represent the starting and finishing positions of the inducing points, respectively. 
	Training epochs are scaled by $10^3$. Best seen in color.}
    \label{fig:ip_evolution}
\end{figure}

Figure \ref{fig:ip_evolution} shows the predictive distribution obtained by SIP, as well as the changes in 
position of the inducing points across epochs (scaled by $1e3$). We see that the inducing points locate themselves covering 
the whole range of the training data. After epoch $2000$ they move very little. This shows that the method is 
able to successfully locate the inducing points to the most convenient position (see Appendix \ref{app:inducing_points_location_extra}). Moreover, the predictive distribution seems good even though we do not fit the parameters of 
$q_\phi(\mathbf{u})$ here.

\begin{figure*}[th]
\setlength{\belowcaptionskip}{-10pt}
\centering
\begin{center}
	\begin{tabular}{ccc}
		\includegraphics[width = 0.3\textwidth] {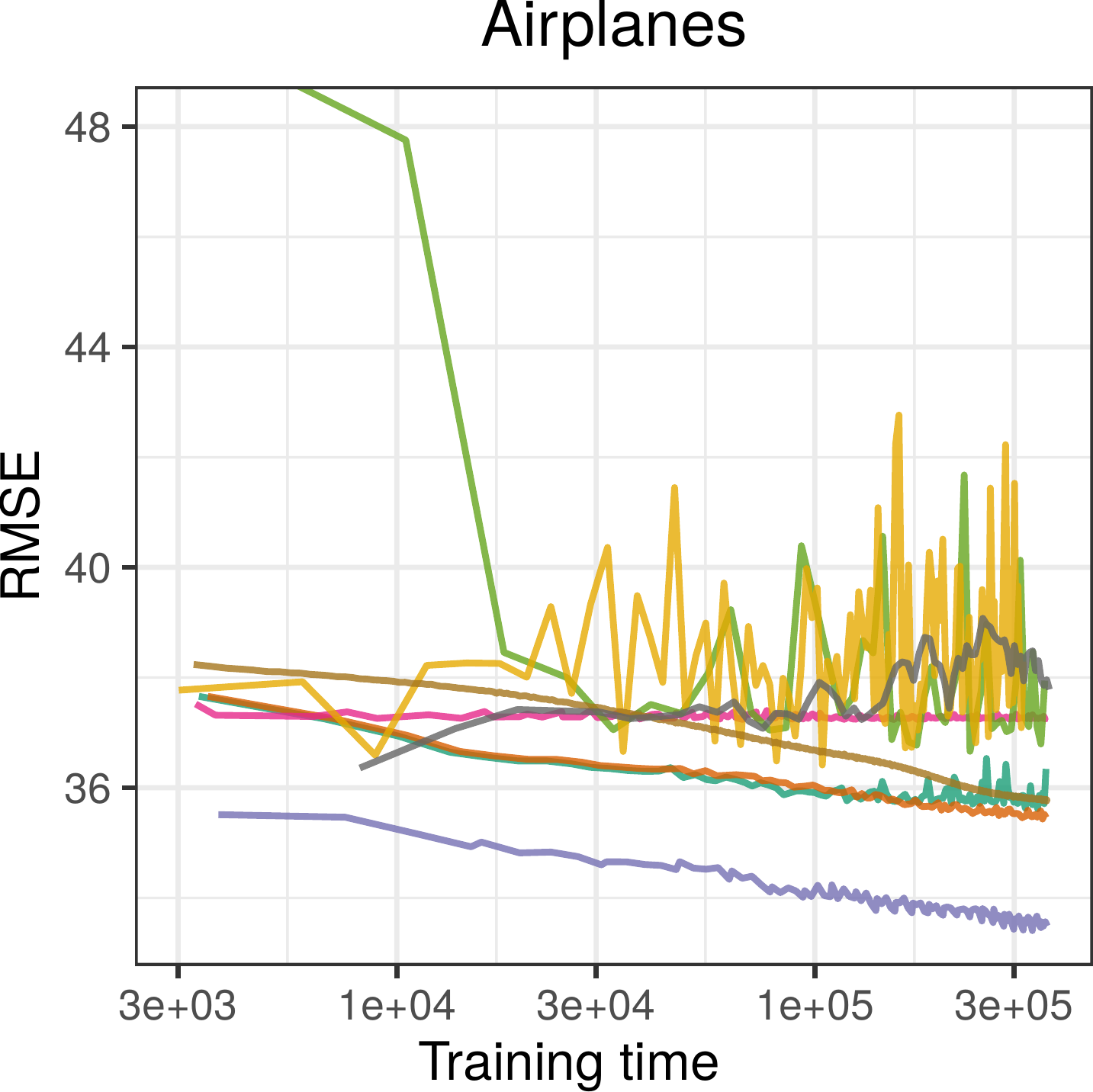} & 
		\includegraphics[width = 0.3\textwidth] {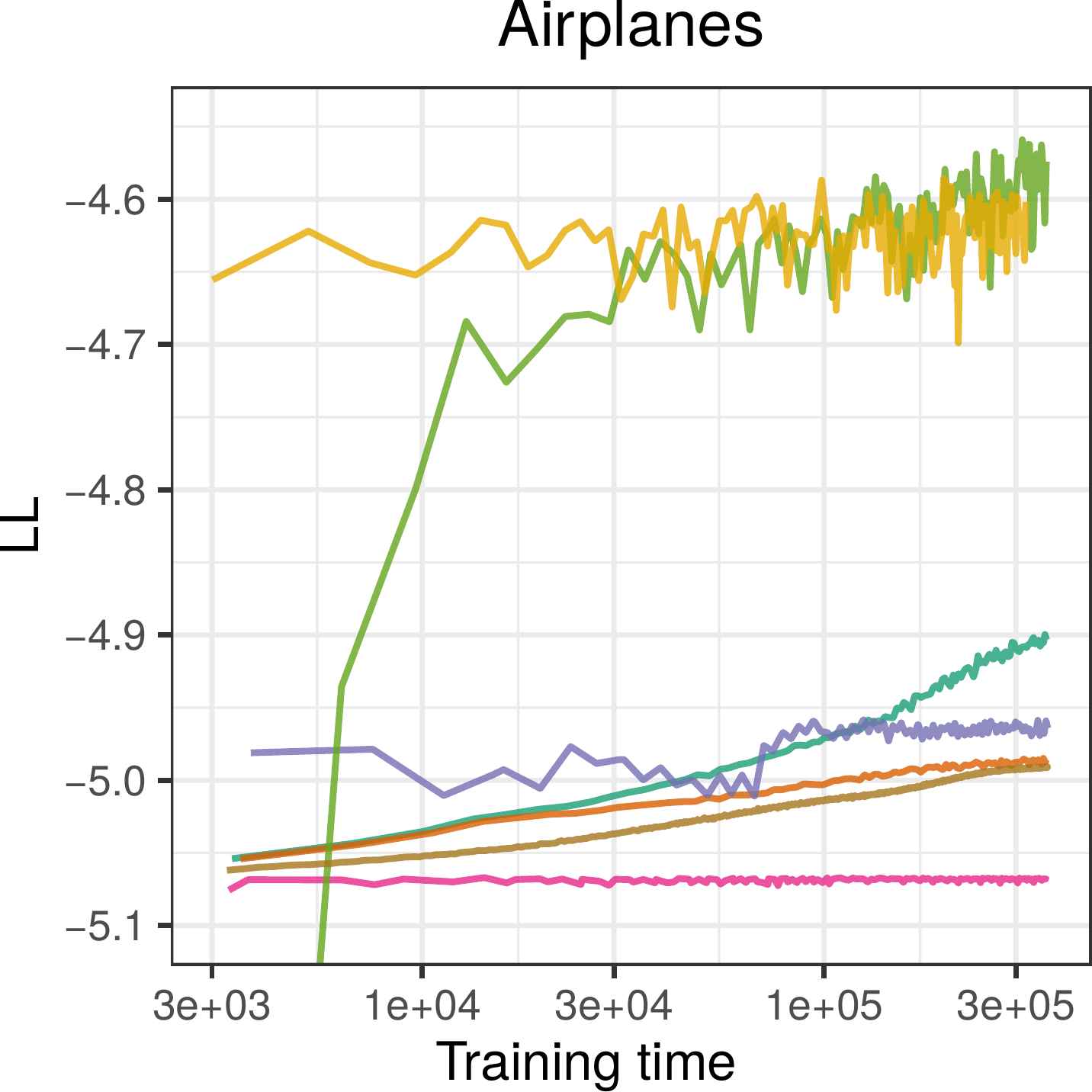} &
		\includegraphics[width = 0.3\textwidth] {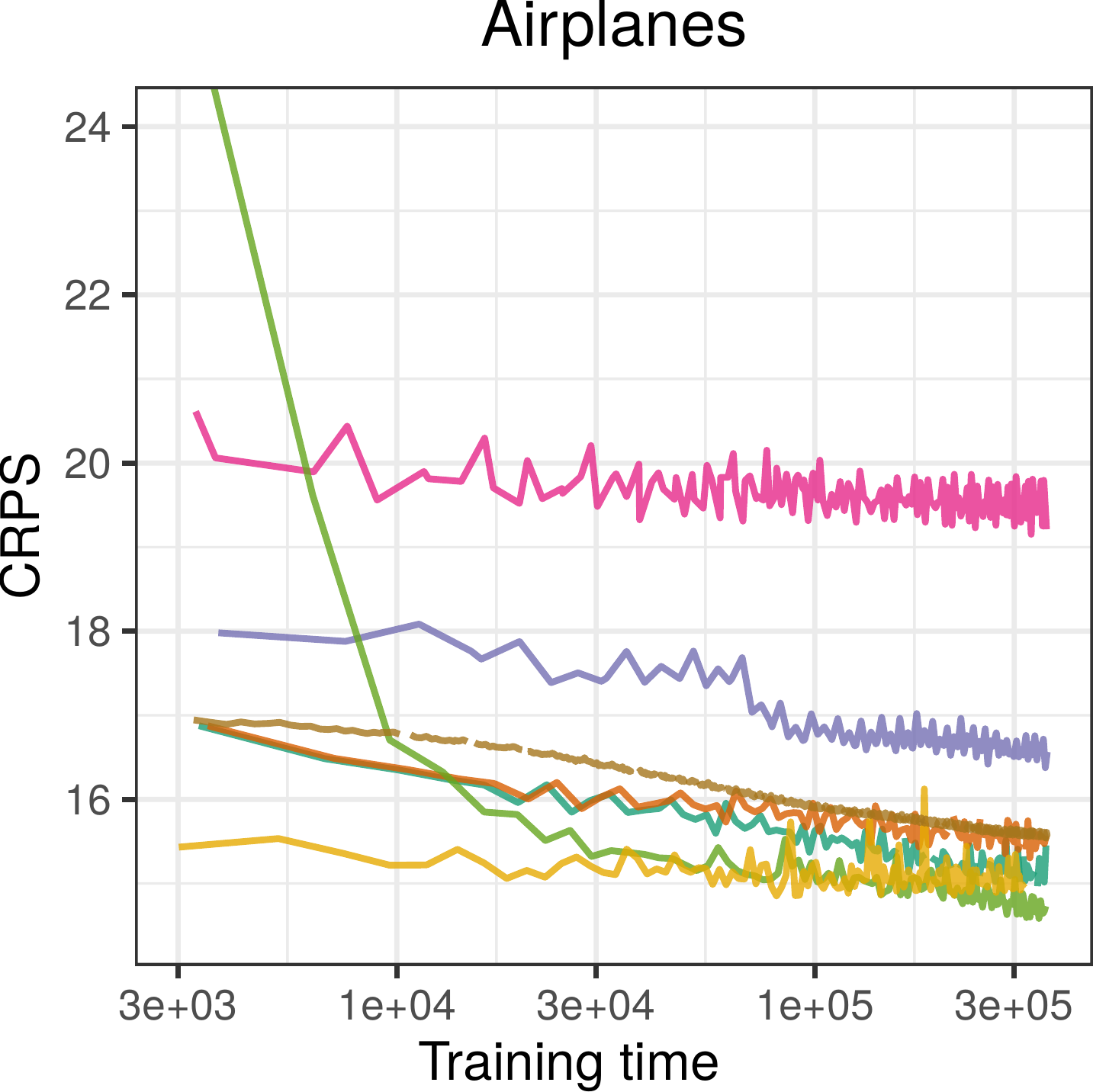} \\
		\includegraphics[width = 0.3\textwidth] {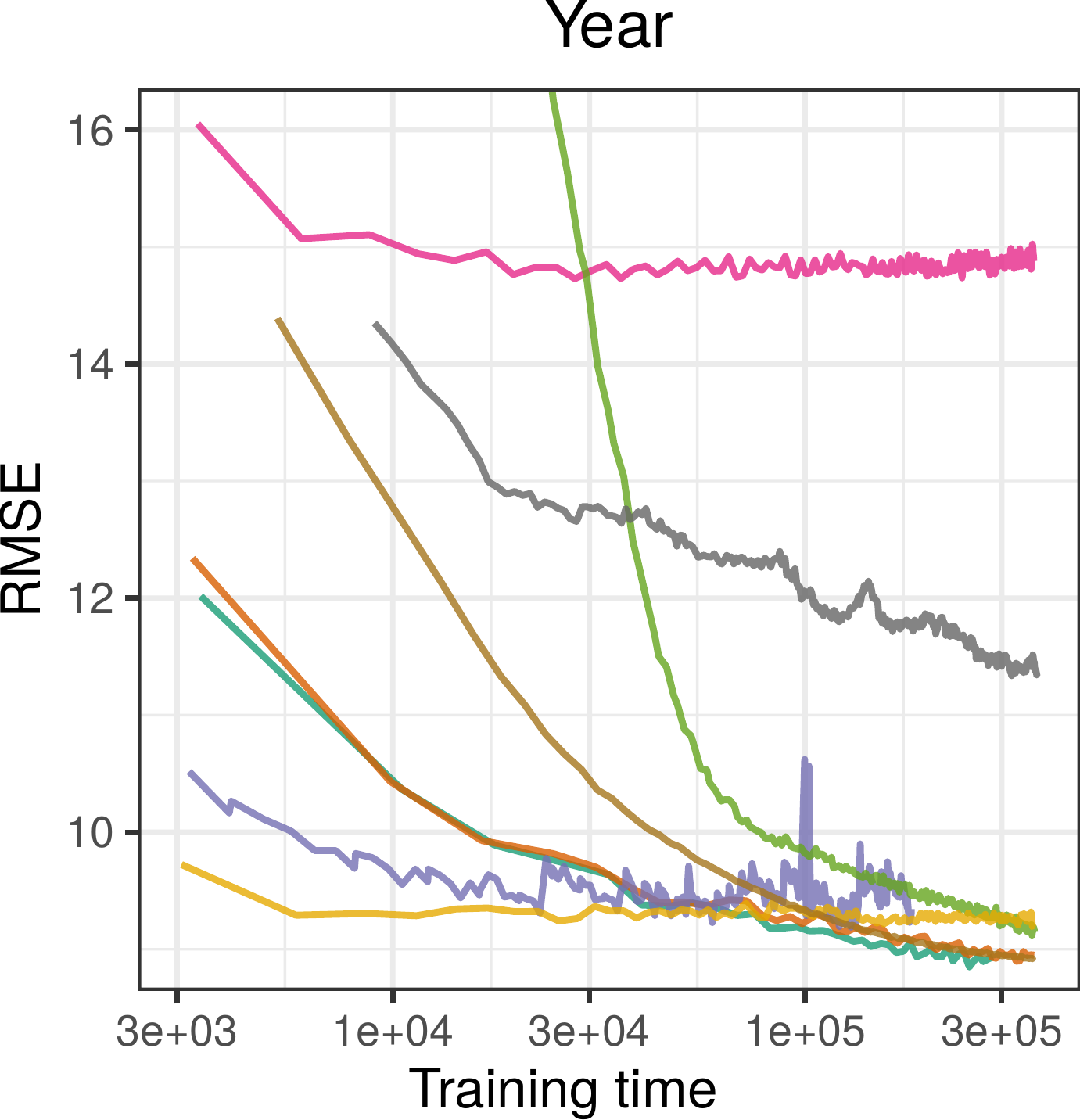} &
		\includegraphics[width = 0.3\textwidth] {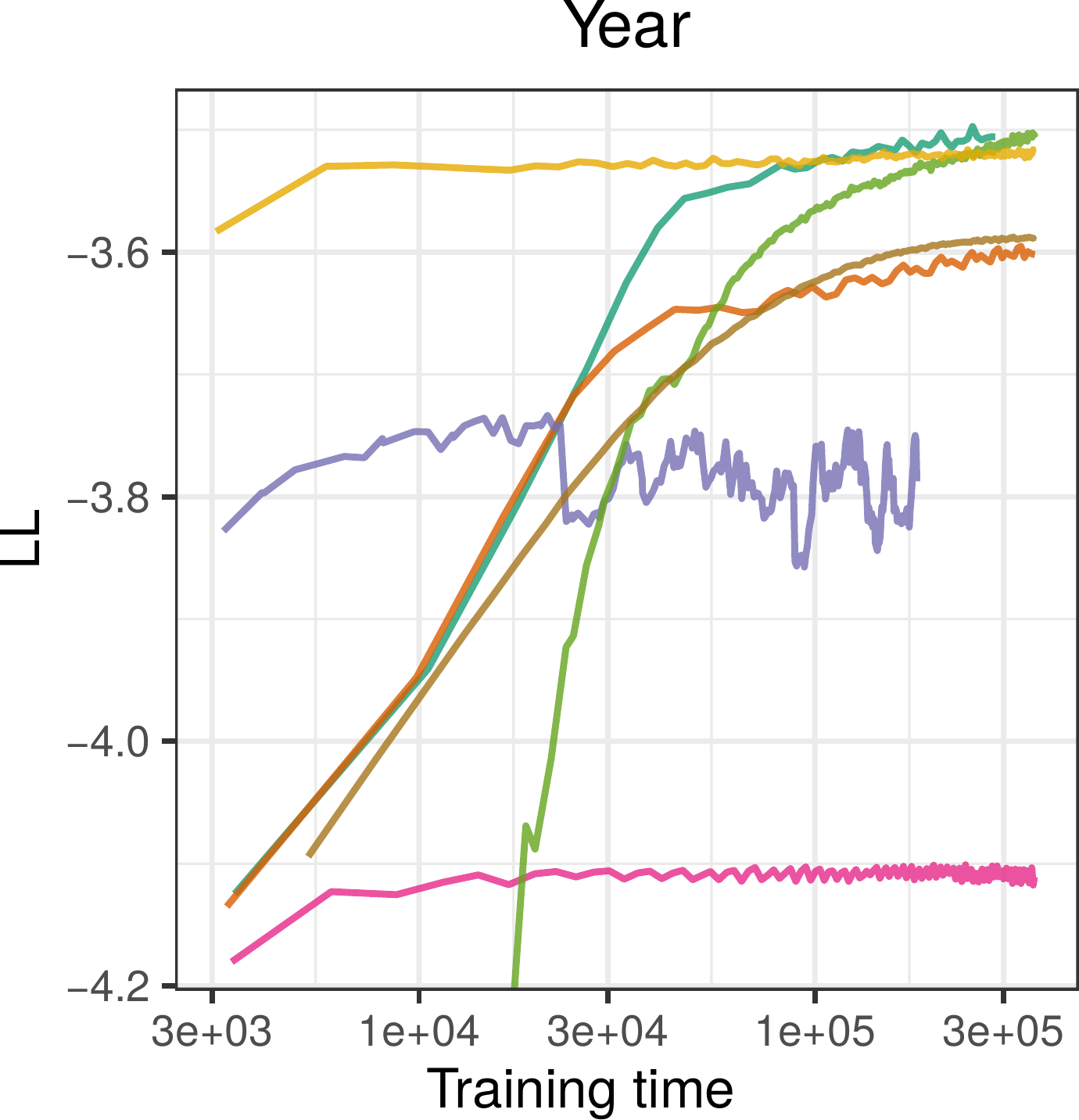} &
		\includegraphics[width = 0.3\textwidth] {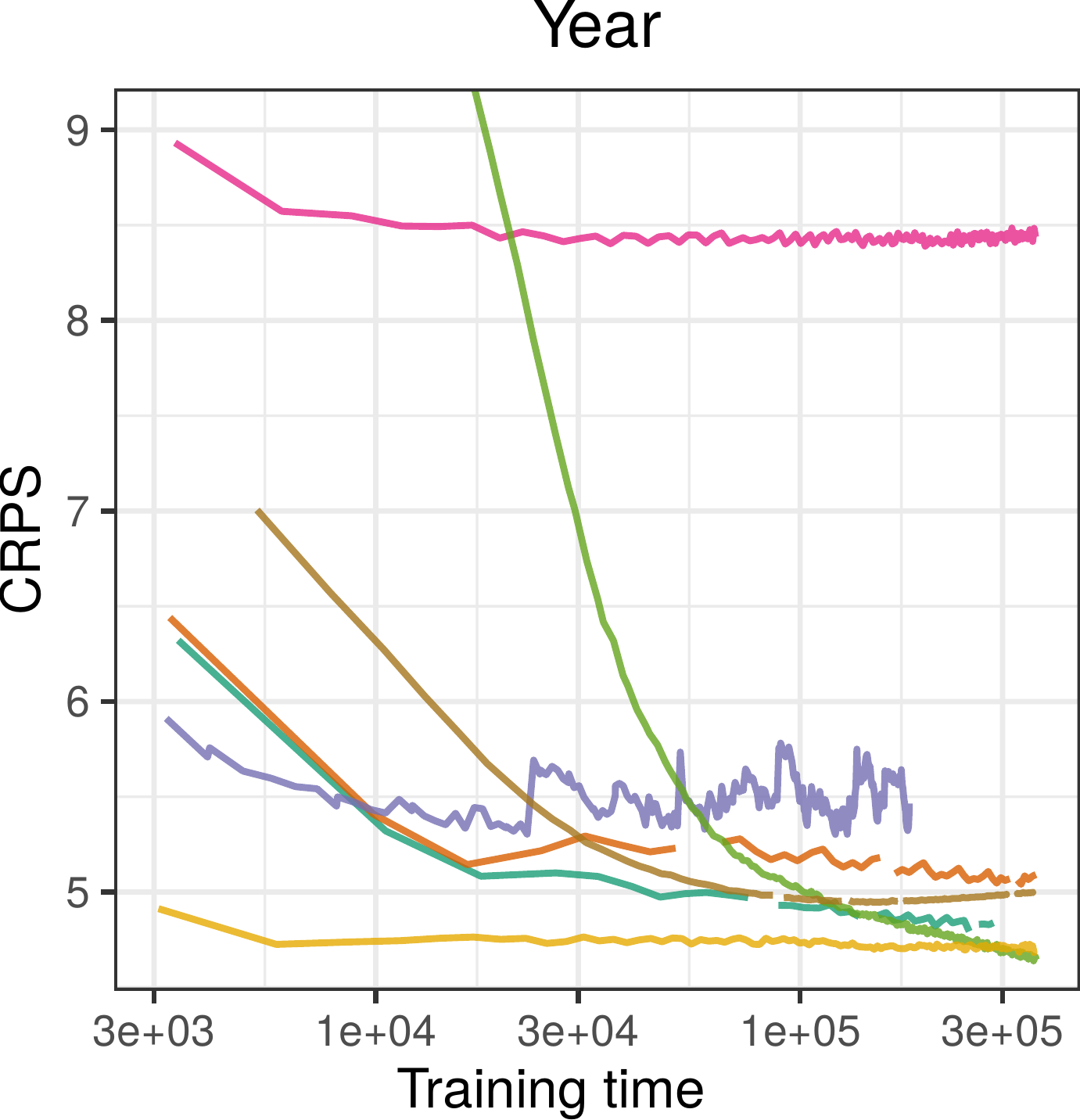} \\
		\multicolumn{3}{c}{\includegraphics[width =0.9\textwidth] {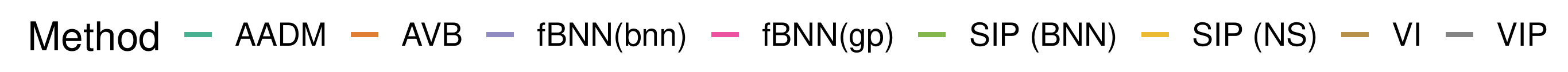}}
	\end{tabular}
 \vspace{-0.15cm}
\caption{\small Performance as a function of training time (in seconds) in a log scale for 
each dataset, method and metric.  Airplanes (first row) and Year (second row). The legend 
of each method is shown at the bottom. Best seen in color.}
\label{fig:large_figure_experiments}
\end{center}
\end{figure*}

\subsection{Regression on UCI Datasets}
\label{subsec:regression_uci_data}

We compare each method and SIP with each prior (\emph{i.e.}, a BNN and a NS) on 
multivariate regression problems from the public UCI dataset repository \cite{Dua2017}. 
We refer to these methods as $\text{SIP}_\text{BNN}$ and $\text{SIP}_\text{NS}$, respectively. 
Following \citet{ma2019variational},
we select 8 datasets: Boston Housing, Concrete, Energy Efficiency, Kin8nm, Naval, Combined Cycle 
Power Plant, Wine and Yatch. We split the data 20 times into train and test with 80\% and 20\% of the 
instances, respectively. The performance metrics employed are the RMSE, the test log-likelihood (LL) and 
the Continuous Ranked Probability Score (CRPS). CRPS is a proper scoring rule that can be 
used as an alternative metric of the accuracy of the predictive distribution \cite{gneiting2007strictly}. 
In the case of fBNN we report results for a BNN prior since it performs better than a GP prior. In SIP we use 
$100$ inducing points. In $\text{SIP}_\text{NS}$ we implemented dropout with a rate of $0.05$. 
SIP, fBNN and AADM use a \textit{warm-up} period on which the KL in \ref{eq:alpha_f-elbo} is 
multiplied by a factor $\beta$ that linearly increases from $0$ to $1$ for the first $20\%$ of the total number of epochs.
Finally, each method is trained until convergence on each dataset. More details on the experimental setup, 
as well as detailed results for each dataset, are provided in Appendix \ref{app:uci_experiments_extra}. 

To get an overall comparison among each method we have ranked them from best to worst for each 
data split: the best method gets rank 1, the second best rank 2, etc. 
These ranks are then then averaged for each metric. Final results are displayed 
in Table \ref{tab:ranking_analysis}, where the average rank across all datasets is 
shown for each method and performance metric (lower is better). We observe that for RMSE, 
$\text{SIP}_{\text{NS}}$ is the best method. Similarly, $\text{SIP}_{\text{BNN}}$ has the smallest average 
rank for test log-likelihood (followed by AADM) and CRPS (followed by VIP). VIP also performs very well in terms of RMSE. 
AADM is also performs well and is a flexible method that can also produce accurate predictive 
distributions \cite{santana2020adversarial}. However, it performs inference in parameter space instead of function space.
This may explain the advantage of SIP over AADM. In terms of CRPS, VIP is the second best method,
probably because this metric is more influenced by the good result in 
the prediction error than the test log-likelihood. In summary, SIP gives competitive 
results with state-of-the-art methods for approximate inference in the context of NNs.

\subsection{Experiments on Large Datasets}
\label{subsec:long_experiments_section}

We also compared each method's performance as a function of the computational training
time. We considered two datasets: (i) \textit{Year Prediction MSD}, 
with $515,345$ instances and $90$ attributes \cite{Dua2017}; (ii) \emph{Airplanes Delay} 
with $2,127,068$ instances and $8$ attributes \cite{santana2020adversarial}. 
A random split of $10,000$ instances is used as the test on which each performance metric is computed for each 
method over training. In SIP we employ $100$ inducing points, with dropout rate $0.05$ for $\text{SIP}_\text{NS}$. 
In VIP and $\text{fBNN}_\text{GP}$, a pre-training procedure on a subset of the data is be carried out to estimate the hyper-parameters. The time for this is added to the total training time.

Figure \ref{fig:large_figure_experiments} shows the results of each method 
for each performance metric as a function of training time. We have removed 
the first $3000$ seconds of time for the sake of visualization. 
The fastest method in terms of performance vs. computational time is $\text{SIP}_\text{NS}$. In the 
case of the test log-likelihood (LL) and CRPS it obtains the best results after a small training time. 
In terms of LL, SIP is the clear winner in Airplanes. In Year, however, AADM also gives good results. 
The same happens in terms of CRPS, where SIP performs best with both priors, \emph{i.e.}, BNN and NS. 
$\text{SIP}_\text{NS}$ is faster than $\text{SIP}_\text{BNN}$ 
by the more efficient sampling approach taking place in the prior in this model, which exploits
a specific architecture that simplifies this task (see \citet{mescheder2017adversarial}). 
The rest of methods perform similarly in general here, with $\text{fBNN}_\text{GP}$ obtaining worse 
results than the other methods and far behind those of $\text{fBNN}_\text{BNN}$. Finally, VIP 
performs bad in general. Here, VIP under-estimates the noise variance producing too confident predictions 
that result in low CRPS and LL (worse than the ones of the other methods, not 
shown in the figures).  Nevertheless, VIP's RMSE is good in general.
SIP and AADM tend to perform better in terms of LL and CRPS than RMSE, which is targeted
for medium values of $\alpha$ \cite{santana2020adversarial} (smaller $\alpha$ values 
target the RMSE). In terms of performance vs. training time, $\text{SIP}_\text{NS}$ is the 
fastest method. However, given enough training time, $\text{SIP}_\text{BNN}$ outperforms the 
other methods in terms of LL and CRPS. In terms of RMSE it  performs similarly to the other 
methods in Year. These experiments confirm again that SIP is competitive with state-of-the-art methods.


\section{Conclusions}
\label{sec:conclusions}

We have proposed SIP for approximate inference.
SIP can be used in several models (\emph{e.g.}, Bayesian NNs and
neural samplers). Moreover, SIP can adjust the prior parameters to
the data. It can also use a flexible implicit process (IP) to approximate the 
posterior distribution. Current methods cannot perform these two tasks simultaneously. 
Importantly, SIP can generate flexible predictive distributions that capture 
complicated patterns in the data such as bimodality or heteroscedasticity 
(see Appendix \ref{app:het_data_experiments}). We have evaluated SIP on several tasks 
for regression. It gives similar and often better results, in terms of several performance 
metrics, than state-of-the-art methods for approximate inference
in the context of Bayesian NNs. SIP is scalable and can be used in datasets with millions of instances.  
This is achieved by a sparse approximation based on inducing points similar to the one 
often used in GPs. Our experiments also show that SIP can 
learn a sensible location for the inducing points. A limitation of SIP is, however, that 
it requires the evaluation of complex conditional distributions. Nevertheless, 
they can be approximated by a GP with the same covariances as 
the prior IP. The covariances can be estimated via Monte Carlo methods.
SIP could have an important societal impact, specially when accurate 
predictive distributions are critical. For example, when the decisions made 
can have an influence on people's life, such as in autonomous vehicles 
\cite{mcallister2017concrete} or in medical diagnosis tools \cite{sajda2006machine}.



\section*{Acknowledgements}

The authors gratefully acknowledge the use of the facilities of Centro de Computacion Cientifica (CCC) at Universidad Autónoma de Madrid. SRS acknowledges the Spanish Ministry of Economy for the FPI SEV-2015-0554-16-4 Ph.D. grant. DHL and SRS also acknowledge financial support from Spanish Plan Nacional I+D+i, PID2019-106827GB-I00 /AEI / 10.13039/501100011033. BZ has been supported by the Programa Atracci\'on de Talento de la Comunidad de Madrid under grant n. 2017- T2/TIC-5455, from the Comunidad de Madrid/UAM “Proyecto de J\'ovenes Investigadores” grant n. SI1/PJI/2019-00294, as well as from Spanish “Proyectos de I+D de Generaci\'on de Conocimiento” via grants PGC2018-096646-A-I00 and PGC2018-095161-B-I00. BZ finally acknowledge the support from Generalitat Valenciana through the plan GenT program (CIDEGENT/2020/055).


\bibliography{bib_SIPs}
\bibliographystyle{icml2022}

\newpage
\appendix
\onecolumn

\section{Gradients analysis}
\label{app:gradient_analysis}

The definition of the original functional ELBO objective for approximate inference, using implicit distribution and a discriminator to estimate the KL divergence term, is the following:
\begin{align}
\label{eq:fELBO_supp}
    \mathcal{L}(q) \: = \: \sum_{i=1}^N \mathds{E}_{q_{\phi,\theta}}[ \log p(y_i|f_i) ] - \mathds{E}_q[T_{\omega^\star}(\mathbf{u})]
\end{align}
where $T_{\omega^\star}(\mathbf{u})$ is the optimal discriminator, approximated by a deep neural network that tells apart samples from $q_\phi(\mathbf{u})$ and $p_\theta(\mathbf{u})$. However, as mentioned in the main text of the article, training the model employing this expression turns out ineffective in terms of updating the prior, since the gradients w.r.t the prior parameters of the second term are either zero or assumed to be negligible \cite{sun2019functional}. To check this argument we performed a few experiments using $\text{SIP}_\text{BNN}$. 

The gradients in the data term of (\ref{eq:fELBO_supp}) w.r.t. any parameter can easily be estimated by automated derivation methods. However, in order to make the comparisons, we need to estimate the gradients in the second term w.r.t. the same parameters as well, and since we are mainly interested in the parameters regarding the prior, they will be $\theta$ and the inducing points' locations $\mathbf{\bar{X}}$. Here, in order to obtain the gradients of the estimated KL term we must first obtain the optimal discriminator value, which results in the most precise estimate of the log-ratio between $q_\phi(\mathbf{u})$ and $p_\theta(\mathbf{u})$. Then, we perturb slightly the value of the selected parameter and retrain the discriminator, obtaining the new estimate for the log-ratio between the distributions and allowing us to estimate the gradient w.r.t. the parameter using finite differences. In both phases, we train the model until convergence to ensure that we have reached $T_{\omega^\star}(\mathbf{u})$.

To compare different contributions to the total gradient of (\ref{eq:fELBO_supp}) we study separately $\theta$ and $\mathbf{\bar{X}}$. In the case of $\theta$, since we use a 2-layered BNN with 50 units in each layer as prior, there are a lot of parameters to choose from. In order to select among them the most relevant ones, we have chosen the 500 parameters that showed the biggest contribution to the gradient in the data term. Afterwards we perform the previous finite differences procedure and compare the gradients obtained in both terms for the same parameters. On the other hand, for the gradients regarding $\mathbf{\bar{X}}$, we estimate the KL gradient for every inducing point, since in this case we are only employing 50 inducing points (as in the synthetic experiments). Therefore, we will be able to compare the gradients of both terms for every inducing point employed.  

\begin{figure}[h!]
\setlength{\belowcaptionskip}{-10pt}
\centering
\begin{center}
	\begin{tabular}{cc}
    \includegraphics[width = 0.48\textwidth] {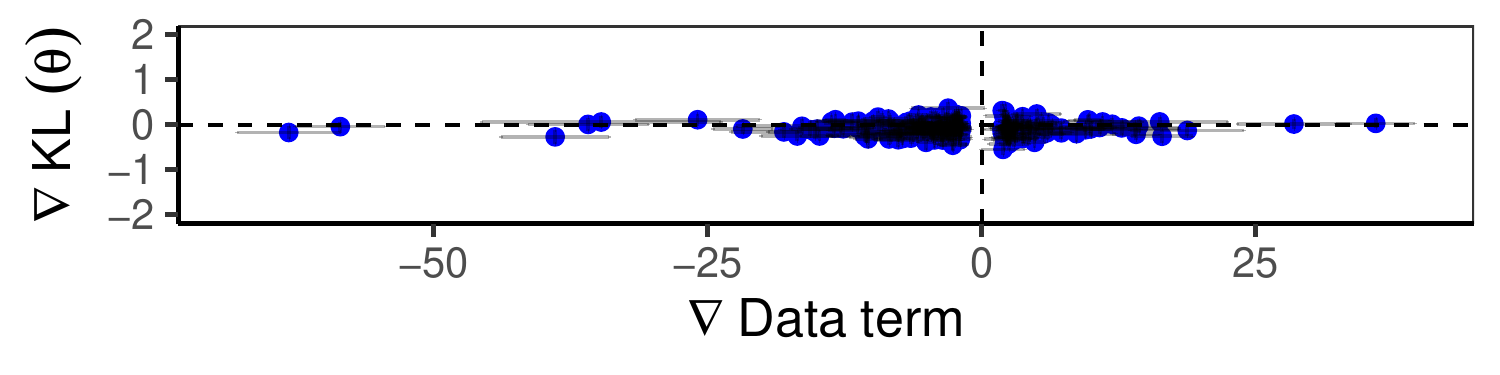} & 
    \includegraphics[width = 0.48\textwidth] {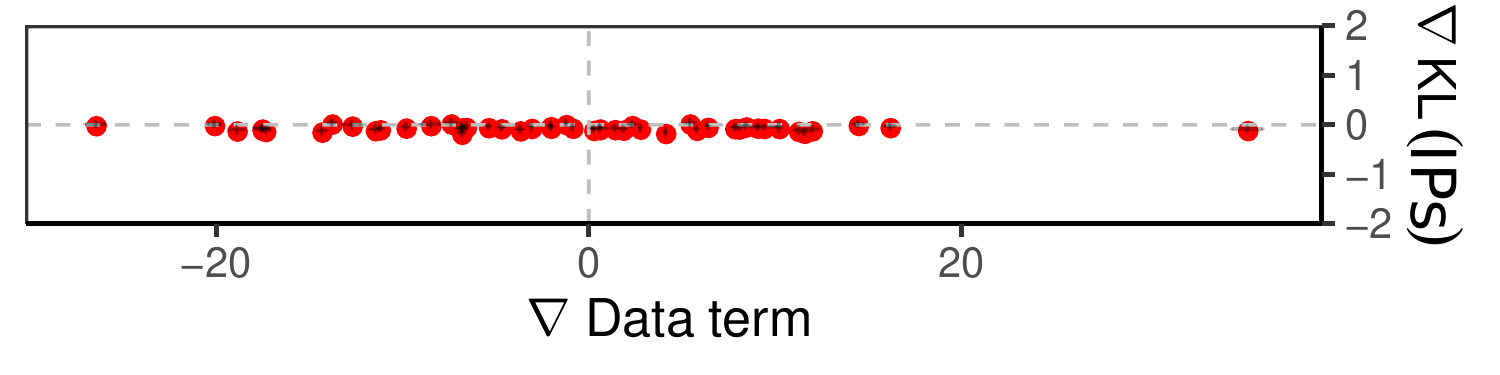} \\
	\end{tabular}
  \caption{{\small Comparison between the gradients of the data term (x-axis) and the two sets of prior parameters in the model, \emph{i.e.} $\theta$ (left, in blue) and the inducing points' locations $\mathbf{\bar{X}}$ (right, in red). The scale of the x-axis is in both cases much bigger than the y-axis for visualization purposes. Error bars are included in both plots, although may not be visible when compared against the size of the points. Best seen in color.}}
  \label{fig:gradients_experiment}
\end{center}
\end{figure}

In Figure \ref{fig:gradients_experiment} we show the comparisons between the gradients of the data term in the f-ELBO and the two possible gradient contributions from the prior parameters, \emph{i.e.} $\theta$ (left image, in blue), and the inducing points' locations (right image, in red). We have included error bars, although specially in the second plot most of them are smaller than the size of the points themselves. In both figures, the x-axis represents the gradient in the data term, and the y-value for each point is the estimated gradient value for the KL term in (\ref{eq:fELBO_supp}) w.r.t. either $\theta$ and $\mathbf{\bar{X}}$. As can be seen, in both cases the x-axis has a wider range than the y-axis by a factor bigger than 10. As we mentioned, when comparing the gradient of the data term against the gradient of the KL term w.r.t. $\theta$ we selected the 500 parameters in $\theta$ that had the largest contributions to the data term, which explains the gap in $x=0$. Nonetheless, we see here that in every case $|y|<1$, which means that $\nabla_\theta \text{KL} \ll \nabla_\theta (\text{Data term})$. Moreover, in the second plot we see the same behavior, meaning that $\nabla_{\bar{{\text{X}}}} \text{KL} \ll \nabla_{\bar{{\text{X}}}} (\text{Data term})$.

\begin{figure}[h!]
    \centering
    \includegraphics[width=0.9\textwidth]{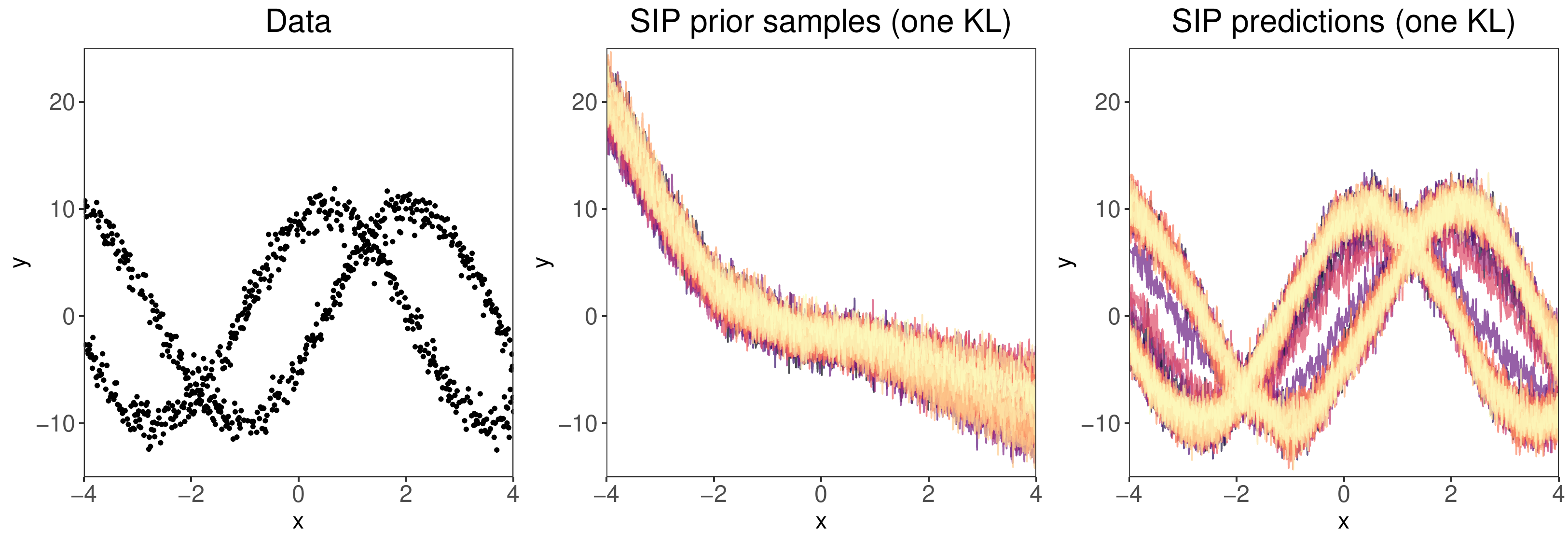}
    \caption{Results of training SIP with the original f-ELBO formulation with only one KL contribution for the bimodal data presented in the main text (\textit{left}). The predictive distribution (\textit{right}) reproduces the bimodal behavior correctly. However, the functions sampled from the final prior model (\textit{center}) do not follow the same behavior of the data.}
    \label{fig:one_kl_comparison}
\end{figure}

SIP can be trained using this original formulation of the objective function. Although the resulting predictive distribution produces good results, these negligible contributions to the gradients makes the system unable to train its prior model. The results from this process, using the $\text{SIP}_{\text{BNN}}$ model,  are shown in Figure \ref{fig:one_kl_comparison}. There, the samples from the prior model can be seen to not follow the behavior of the original data, while the predictive distribution is capable of reproducing the original bimodal behavior. This means that the prior is not being properly trained, although the posterior model is flexible enough to compensate for this fact and accurately follow the data behavior.

From these results we conclude that the contribution to the gradient of the original objective function from the prior parameters is much less relevant than the one coming from the data term. Therefore, we need to introduce the new $\mathcal{L}^\star(q)$ objective, which allows us to train simultaneously both the prior and posterior parameters. 

\section{Adaptive contrast for SIP}
\label{app:adaptive_contrast}

Adaptive contrast (AC) is introduced by \cite{mescheder2017adversarial} as an attempt to improve the accuracy of the estimation of the log ratio between distributions in the KL. This estimation will follow from the result of the auxiliary discriminator problem, whose optima is $T_{\omega^\star}$. However, since the distributions being compared are usually very different, the discriminator has no problem telling apart samples from each of them, which does not encourage for an optimal fit of its parameters. Therefore, to make this task harder, an extra Gaussian distribution is introduced in \cite{mescheder2017adversarial}. 
In our case, since the KL distribution is estimated between two IPs, we need to reformulate the original AC to accommodate for this fact. This will imply extending the original AC to use two discriminators instead of one. Let us define two Gaussian distributions $\bar{q}$ and $\bar{p}$ with the same moments as the samples from both $q$ and $p$ (our IP approximating posterior and IP prior)
\begin{equation}
    \bar{q} \sim \mathcal{N}(\mu(q), \sigma^2(q)), \qquad \bar{p} \sim \mathcal{N}(\mu(p), \sigma^2(p))
\end{equation}
with $\mu(\cdot)$ and $\sigma^2(\cdot)$ as the mean and variance across samples from $q$ or $p$. Using this, we rewrite the two-KL part of the f-ELBO objective as the following:
\begin{align}
    \text{KL}(q|p) + \text{KL}(p|q) &= \mathds{E}_q\left[\log \frac{{\color{red} q} \bar{q} {\color{blue} \bar{p}}}{{\color{blue}p} {\color{red} \bar{q}} \bar{p}}\right] + \mathds{E}_p\left[\log \frac{{\color{blue}p} \bar{p} {\color{red} \bar{q}}}{{\color{red}q} {\color{blue} \bar{p}} \bar{q} }\right] \nonumber \\ 
    &= \mathds{E}_q\left[ \log {\color{red}\frac{q}{\bar{q}}}\right] + \mathds{E}_q\left[ \log {\color{blue}\frac{\bar{p}}{p}}\right] +  \mathds{E}_q\left[\log \frac{\bar{q}}{\bar{p}} \right] \nonumber \\ 
    & \qquad + \mathds{E}_p\left[\log {\color{blue}\frac{p}{\bar{p}}}\right] + \mathds{E}_p\left[\log {\color{red}\frac{\bar{q}}{q}}\right] + \mathds{E}_p\left[\log \frac{\bar{p}}{\bar{q}}\right] \nonumber \\
    &= \mathds{E}_q [T({\color{red}q,\bar{q}}) - T({\color{blue}p,\bar{p}})] + \mathds{E}_p [T({\color{blue}p,\bar{p}}) - T({\color{red}q,\bar{q}})] \nonumber \\ 
    & \qquad + \mathds{E}_q\left[ \log\frac{\bar{q}}{\bar{p}} \right] + \mathds{E}_p\left[ \log \frac{\bar{p}}{\bar{q}} \right].
\end{align}

Now we will have to employ \textbf{two} discriminators, one to separate samples from $q$ and $ \bar{q}$ ($T(q, \bar{q})$), and another to separate samples from $p$ and $\bar{p}$ ($T(p, \bar{p})$). The two last contributions to this expression are given by the log-ratio of two Gaussian distributions with given mean and variances, and are thus tractable and have a closed-form solution.

In practice, both discriminators needed here are estimated once with the expected value w.r.t. samples from $q$ and from $p$. To help with this task, one could standardize the samples from $q$ and $p$ in beforehand, following the description in \cite{mescheder2017adversarial, santana2020adversarial}
\begin{align}
    \text{KL}(q|p) + \text{KL}(p|q) &= \text{KL}(q_0|p_q) + \text{KL}(p_0|q_p),
\end{align}
where 
\begin{equation}
    q_0 = \frac{q - \mu(q)}{\sigma(q)}, \qquad p_0 = \frac{p - \mu(p)}{\sigma(p)}, \qquad q_p = \frac{q - \mu(p)}{\sigma(p)}, \qquad p_q = \frac{p - \mu(q)}{\sigma(q)}.
\end{equation}
This simplifies the involved calculations for the discriminators involved, since now several terms employ standardized Gaussian distributions. 

We opt to include the description of AC here for the sake of completeness, although in the experiments conducted in the main text, we found AC produced overfitting in the training of the prior. This would then require regularization techniques such as dropout or the \textit{llh-regularization} used in VIP \cite{ma2019variational}. We have tested these models as well, and thus far we have seen they also provide good performance. However, we think the added difficulties of training an extra discriminator and the need for a regularization term surpasses the benefits for this extra additions to the original SIP model.


\section{HMC implementation and remarks}
Ideally, the ground truth for Bayesian inference is provided by Markov Chain Monte Carlo (MCMC) sampling methods, where a Markov Chain is designed to converge to the true underlying distribution after some -potentially very large- number of steps. In particular, the Hybrid -or Hamilton- Monte Carlo (HMC) method \cite{neal2011mcmc} is one of the most popular methods nowadays. The HMC method avoids the inefficient random-walk behavior of other MCMC methods by using the gradient information of the target distribution. This is done by solving the Hamilton equations, where a set of latent "momentum" variables are introduced and associated in a one-to-one correspondence to the original variables making up the parameter space of the model.  Specific implementation of HMC involves choosing a concrete choice for the "kinetic energy" of the Hamiltonian, which is equivalent to the choice for the distribution of the momentum variables, as well as a particular algorithm for solving the Hamilton equations. A common practice for the former which has been adopted in this work, is to assume a multivariate standard Gaussian. On the other hand, we also follow the standard practice and solve the Hamiltonian equations with the so-called leapfrog algorithm, which solves the system of differential equations along a trajectory consisting of $L$ integration steps of size $\epsilon$. For the datasets below, we have found that L=25 and $\epsilon=5\cdot 10^{-5}$ give sufficiently good results. Changing the covariance matrix of the momentum variables (e.g. by considering a more general diagonal matrix) gives roughly the same overall performance after properly tuning the above leapfrog parameters. In this work we have found that, for all the problems in consideration, a Markov chain of 10,000 steps was enough to both attain the equilibrium distribution.  

In practice, however, HMC may not be able to capture the true underlying distribution. Beyond the obvious limitation of computational resources, which could prevent the Markov chain to converge as desired or to have a sufficiently large number of truly independent samples, there is another reason typically overlooked by the common lore. That is, in order for the HMC to attain the true target distribution the assumed model should be the correct one. For example, in the case of the bimodal dataset analyzed below, if the proposed joint distribution of the parameters and the data is unimodal, then the HMC will never converge to the desired distribution, even in the limit of infinite number of Markov chain steps. An analogous situation is encountered with the heterocedastic dataset shown above. In general, if the data comes from a distribution having features not captured by the proposed model, the MCMC will not deliver the desired outcome.  In Figure 1 of the main document we observe the HMC applied to the bimodal dataset. As already commented there, it does not capture the bimodality, simply because the model assumed a Gaussian prior and a Gaussian likelihood. This cannot explain the bimodal data. In a similar way, in Figure \ref{fig:prior_and_predictive_heteroc} we observe that the HMC is unable to capture the heteroscedasticity of the data, because the model assumes homocedasticity instead.

\section{Extra synthetic data comparisons}
\label{app:extra_synth_examples}

\subsection{Alternative setups for bimodal data}
\label{subsec:alternative_bimodal_setups}

The experiments conducted with synthetic data in the main text were all performed using the same prior model, a BNN with 2 layers with 50 units per layer. However, to produce those comparisons we have made changes on both methods that could affect their final results: the original code for VIP has a regularization term that was removed for the comparisons there, helping both the prior and the predictive distribution to represent heterocedastic behavior \cite{ma2019variational}. On the other hand, fBNN can employ a GP prior (or a sparse GP), which would have to be trained in beforehand \cite{sun2019functional}. For the sake of completeness, we have run these tests as well with the same bimodal data being used in the main text \cite{depeweg2016learning,santana2020adversarial}. 
\begin{figure}[h!]
    \centering
    \includegraphics[width=0.75\textwidth]{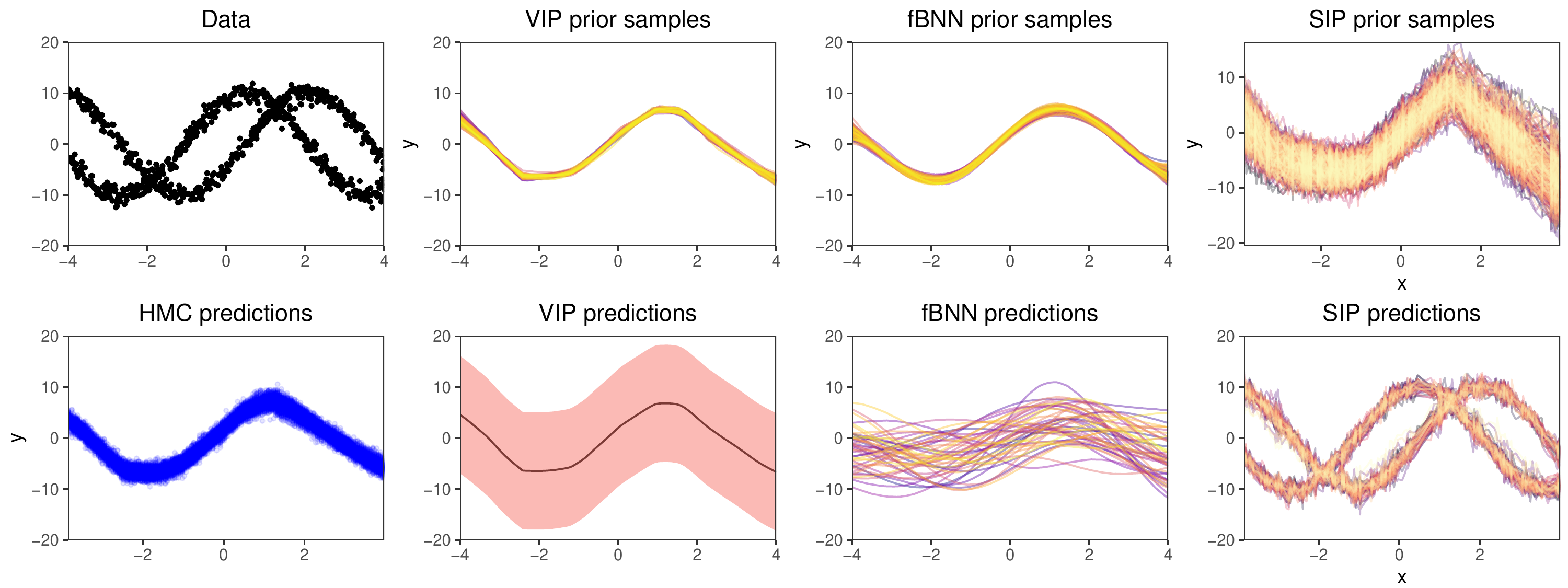}
    \caption{{\small Comparison between methods using bimodal data and alternative setups for VIP and fBNN: VIP includes the regularization term as in the UCI regression experiments, and fBNN uses the GP prior. The first column represents the data distribution (first row, in black) and the HMC predictive distribution (second row, in blue). The remaining figures represent the prior samples (first row) and the predictive distribution (second row) for each method, namely $\text{VIP}^*$ (with regularizer), $\text{fBNN}_\text{GP}$ and $\text{SIP}_\text{BNN}$. For the predictive distribution of $\text{VIP}^*$ the black line represents the predictive mean and the shaded area represents the $\pm 2\sigma$ region. For comparison w.r.t. the original image in the main text, the y-axis range has been increased here so that every plot could be easily seen. Best seen in color.}}
    \label{fig:extra_bimodal_comparison}
\end{figure}

In Figure \ref{fig:extra_bimodal_comparison} we can see the comparisons between methods using these alternative setups, namely VIP with the regularization term (denoted as $\text{VIP}^*$) and fBNN with the GP prior ($\text{fBNN}_\text{GP}$). As in the figure in the main text, the first column represents the training data (first row, in black), and the predictive distribution obtained from applying HMC to the same system on which SIP is implemented, sharing the trained prior parameters (second row, in blue). The other figures represent both the prior samples (first row) or the predictive distribution (second row) for $\text{VIP}^*$, $\text{fBNN}_\text{GP}$ and $\text{SIP}_\text{BNN}$ respectively. In the case of $\text{fBNN}_\text{GP}$ we see that the prior is now being trained (although that needs to happen separately from the rest of the model), resulting in samples that fit better the behavior of the data here. The predictive distribution follows the mean of the data as it did in the case of $\text{fBNN}_\text{BNN}$, and although the noise is slightly smaller in this case, it is not able to reproduce the original bimodality. However, an important detail when comparing these figures against the ones present in the main text is that the y axis range has been increased here w.r.t. the one used in the original one, which may make these figures seem more concentrated towards the mean of the data. 

On the other hand, for $\text{VIP}^*$ we can clearly see that both figures are now not showing any of the heterocedastic behavior they had in the main text. In this case, the prior, as well as the predictive distribution, fit closely the mean of the data, and the regions of $\pm 2 \sigma$ in the latter have a fixed width, unlike in the original figure. Thus, removing the regularizer has helped the method to fit a more expressive prior and to obtain better predictive distribution samples, since introducing it hinders the expressiveness of the model overall. However, since this is the original setup the authors in \cite{ma2019variational} propose, this is the one we employed in both the UCI multivariate regression problems as well as in the convergence experiments. Finally, the good performance of the method regarding the RMSE metric can be explained by its focus on adjusting the predictions to the mean of the training data.

\subsection{Heterocedastic data}
\label{app:het_data_experiments}

To test the ability of SIP to reproduce other complex features present in the training data, we have employed as well an heterocedastic dataset \cite{depeweg2016learning, santana2020adversarial}. We constructed this dataset by uniformly sampling $1000$ values for $x$ between $[-4, 4]$ and then obtaining $y$ for each sample as
\begin{equation}
    y = 7\sin (x) + \epsilon \sin(x) + 10, \:\:\:\:\: \epsilon \sim \mathcal{N}(0, \sigma = 2).
\end{equation}
To perform these experiments we employ the same setup as the one in the main text for the bimodal problem.

\begin{figure}[h!]
    \centering
    \includegraphics[width=0.75\textwidth]{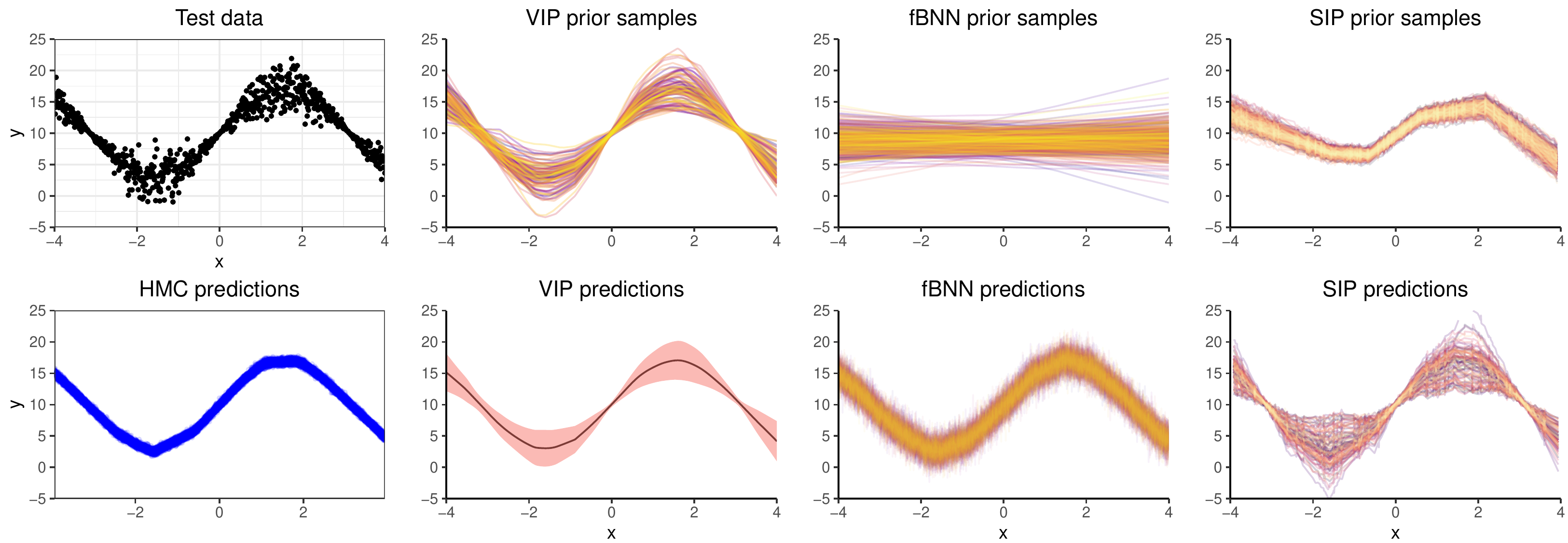}
    \caption{{\small Comparison between methods using heterocedastic data. The first column represents the data distribution (first row, in black) and the HMC predictive distribution (second row, in blue). The remaining figures represent the prior samples (first row) and the predictive distribution (second row) for each method, namely $\text{VIP}$, $\text{fBNN}_\text{BNN}$ and $\text{SIP}_\text{BNN}$. For the predictive distribution of $\text{VIP}$ the line represents the predictive mean and the shaded area represents the $\pm 2\sigma$ region. Best seen in color.}}
    \label{fig:prior_and_predictive_heteroc}
\end{figure}

In Figure \ref{fig:prior_and_predictive_heteroc} we present the results of training the same models we used in the main text for the bimodal dataset (HMC, VIP, $\text{fBNN}_\text{BNN}$ and $\text{SIP}_\text{BNN}$) but with the new heterocedastic synthetic data (represented in the first column, first row, in black). Here, as was the case earlier, HMC (first column, second row, in blue) is unable to reproduce the heterocedastic behavior. Even though it uses the trained prior from SIP, since the selected NN model is not originally capable of reproducing this behavior, HMC is unable to reproduce it either: the prior does not behave in an heterocedastic manner, and neither does the likelihood, resulting in a non-heterocedastic predictive distribution. For the rest of the methods we have the prior samples (first row) and the predictive distribution samples (second row) to compare against them. In the case of VIP (second column), opposite to what is done in Section \ref{subsec:alternative_bimodal_setups}, the regularization term here is turned off again, thus obtaining heterocedastic behavior both in the prior and predictive distributions. If the regularization term is turned on, however, the heteroscedasticity here is lost, as happened in Section \ref{subsec:alternative_bimodal_setups}. For fBNN (third column) we see that the BNN prior is not being trained. Its predictive distribution follows the original data mean closely although without any heteroscedasticity. Finally, in SIP (last column) we can see that the prior behaves somewhat similarly to the original data distribution mean, as it did in the bimodal case. In this case, SIP's predictive distribution also closely follows that of the original data, with heterocedastic behavior as well. In this case, if VIP uses the regularization term, SIP is the only method able to reproduce this behavior, resulting in a better final predictive distribution than the one provided by HMC. This implies what we already mentioned in the main text: when selecting the wrong model, HMC may be unable to reproduce important features from the data that SIP is capable of capturing in its predictive distribution, making it much less susceptible to implicit bias errors caused by choosing the incorrect model for a given problem.

\subsection{Inducing points evolution}
\label{app:inducing_points_location_extra}
To complement the analysis of the evolution of the location of the inducing points in the main text, we conducted an extra experiment using an alternative dataset. This dataset is generated by sampling uniformly $1000$ values for $x$ between $[-4,5]$, and then obtaining $y$ from the following definition:
\begin{equation}
    y = \begin{cases} 
    10 + \epsilon, & x < 0 \\ 
    10(1 + \sin (x)) + \epsilon, & x \geq 0,
    \end{cases}
\end{equation}
with $\epsilon \sim \mathcal{N}(0, \sigma = 2)$. The piece-wise definition of $y$ is constructed so that $y$ is continuous but with a non-defined derivative in $x=0$. We train SIP with this dataset until it converges, and register the location of the inducing points for each epoch of training ($2000$ here). We employ 50 inducing points here for visualization purposes.

\begin{figure}[h!]
    \centering
    \includegraphics{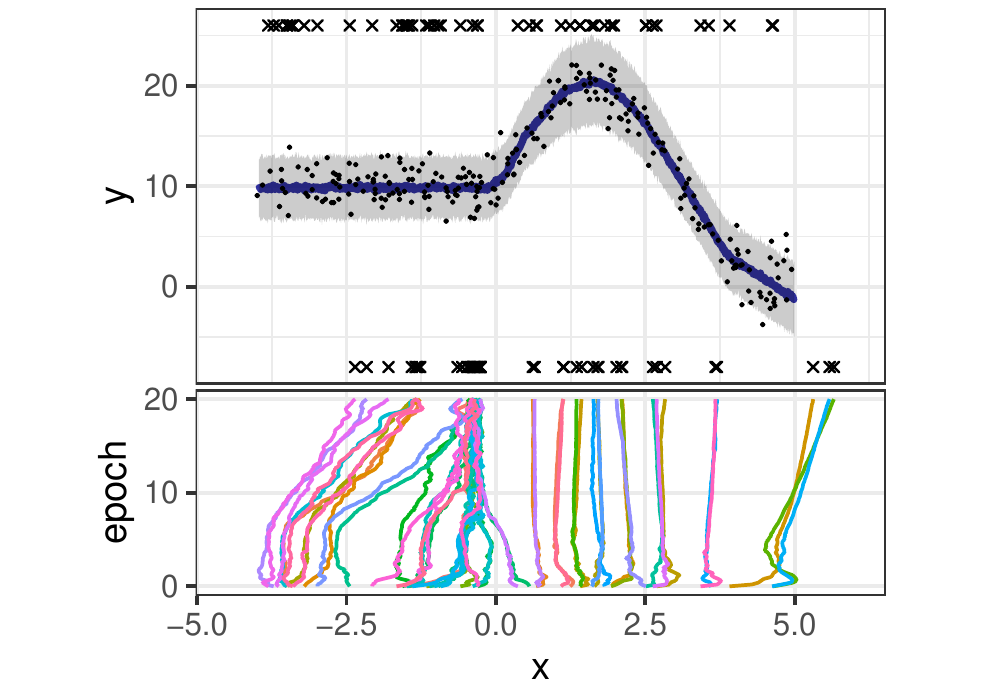}
    \caption{Results of the SIP method applied to the piece-wise defined synthetic dataset. The first figure represents the samples from the predictive distribution of the method, with the mean as the blue line and the training points as black dots. The crosses at the top and bottom represent the starting and finishing locations of the inducing points. The second figure represent the locations of the inducing points for every training epoch in the y-axis. The number of epochs is scaled by $1e2$, ranging from $0$ to $2000$. Best seen in color.}
    \label{fig:ip_evolution_composite}
\end{figure}

In Figure \ref{fig:ip_evolution_composite} we see the results of training SIP with this alternative dataset, using the same representation we employed in the experiment included in the main text: the first part of the figure contains the original data (black dots), the predictive distribution samples (with the mean as a blue line), and the initial and final location of the inducing points represented as crosses both in the top and bottom of the figure respectively. The second plot represents the location of each inducing point for each epoch in the y-axis (scaled by a factor of $100$, ranging from $0$ to $2000$). Moreover, in this case, we have set the starting positions of the 50 inducing points in random locations throughout the whole range of the training set. In the figure, we see that the predictive distribution seems to follow closely that of the original data. However, the most interesting fact here is that the inducing points that started in the region where the data is constant ($x<0$) seem to have a tendency to concentrate around the region of $x=0$, moving right to the matching point between both pieces of the function of $y$. Due to the simple nature of the data before this point, the model focuses the inducing points originally positioned in this region and places them closer to $x=0$, helping it when modeling the change in behavior between the two parts of $y$. On the other side, for the $x>0$ region, the inducing points are distributed more homogeneously to model the sine behavior. Furthermore, the right-most points seem to stray away from the region of the dataset, which is caused by employing far more inducing points than needed in such a simple dataset. This has been done for illustration purposes only, which also causes the overlapping of points across the dataset when there are more than what are needed in the same region. From all of this, we conclude that SIP seems able to distinguish between simpler and more complex regions of the data, and employs its resources effectively to model both at the same time. 

\section{Further details for the convergence experiments}

The convergence experiments have been conducted employing the same computational resources for each method: Each of them employed on its own 2 CPU Intel(R) Xeon(R) Gold 5218 CPU at 2.30GHz (16 cores, 22 Mb L3 cache), [32 cores in total], with 
192 GB RAM at 2,4 GHz. In the experiments, only the training time is reported. The same setup is used for both experiments (\textit{Airplanes delay} and \textit{Year Prediction MSD} datasets). 

To obtain the results shown in the main text, we have chosen to execute VIP without the regularization term described in the supplementary material for the article \cite{ma2019variational}. After conducting several tests, we have concluded that the effect of such regularization term is negligible for these experiments given the size of the datasets. 

Finally, the fast rate of convergence shown by $\text{SIP}_{\text{NS}}$ can be explained by its architecture. The NS follows the system described in \cite{mescheder2017adversarial} to be able to efficiently sample the moments of the approximate distribution for all data points in each mini-batch, which is a time-consuming task in other cases. This allows SIP to be much faster when employing the NS model for the prior, in comparison to what is shown for $\text{SIP}_{\text{BNN}}$, although given enough time the latter usually surpasses the performance of $\text{SIP}_{\text{NS}}$.

\section{Regression results for UCI datasets}
\label{app:uci_experiments_extra}

We include here the mean results for the UCI datasets for each method and metric. Each method is trained on 20 different random train/test splits of the dataset, and the values reported here represent the mean performance of each method in each dataset for the given metric in each case. The \textit{Continuous Ranked Probability Score} (CRPS) is estimated using the \textit{properscoring}\footnote{\url{https://pypi.org/project/properscoring/}} package, both in cases where the predictive distribution is implicit (\emph{e.g.} SIP, AVB and AADM), or when it has an explicit form (\emph{e.g.} VI, VIP). For each dataset, the winning method is highlighted in bold while the one in second place is marked in red (in LL, the higher the better; in RMSE and CRPS, the lower the better) This is done as well for the rankings. The final ranking analysis follows the description of the main text, placing SIP as the leading method for each metric, with the BNN prior for LL and CRPS, and for RMSE with the NS prior.


\vspace*{-0.2cm}

\begin{table}[h!]
\centering
\begin{minipage}[c]{\columnwidth}
\caption{{\normalsize LL results in UCI datasets}} 
\tiny
\noindent
\centering
\begin{tabular}{c|lll|ll|ll}
\hline
\textit{Dataset} & BBB & AVB & $\text{AADM}$ & $\text{fBNN}_\text{BNN}$ & VIP & $\text{SIP}_\text{BNN}$ & $\text{SIP}_\text{NS}$\\
\hline
boston & -2.816$ \pm $0.183 & {\color{red}-2.409}$ \pm ${\color{red}0.153} & $\mathbf{-2.348}$$ \pm $$\mathbf{0.151}$ & -3.193$ \pm $0.687 & -2.559$ \pm $0.409 & -2.51$ \pm $0.321 & -2.595$ \pm $0.521\\
ccpp & -2.844$ \pm $0.0449 & -2.819$ \pm $0.0499 & -2.816$ \pm $0.0495 & -2.789$ \pm $0.0805 & -4.018$ \pm $0.965 & $\mathbf{-2.761}$$ \pm $$\mathbf{0.0507}$ & {\color{red}-2.775}$ \pm ${\color{red}0.0494}\\
concrete & -3.331$ \pm $0.0567 & -2.976$ \pm $0.0928 & $\mathbf{-2.896}$$ \pm $$\mathbf{0.102}$ & -3.514$ \pm $0.491 & -3.208$ \pm $1.04 & -3.364$ \pm $0.424 & {\color{red}-2.934}$ \pm ${\color{red}0.165}\\
ee & -2.155$ \pm $0.0733 & -1.678$ \pm $0.0825 & -1.239$ \pm $0.101 & -1.284$ \pm $0.0485 & $\mathbf{-0.995}$$ \pm $$\mathbf{0.146}$ & {\color{red}-1.002}$ \pm ${\color{red}0.292} & -1.106$ \pm $0.0888\\
wine & -0.979$ \pm $0.0481 & {\color{red}-0.955}$ \pm ${\color{red}0.0479} & $\mathbf{-0.949}$$ \pm $$\mathbf{0.0476}$ & -19.77$ \pm $4.32 & -0.992$ \pm $0.0691 & -0.972$ \pm $0.0683 & -1.149$ \pm $0.168\\
yatch & -2.588$ \pm $0.0968 & -1.742$ \pm $0.085 & -1.559$ \pm $0.157 & -2.114$ \pm $0.0299 & $\mathbf{-0.041}$$ \pm $$\mathbf{0.511}$ & {\color{red}-0.207}$ \pm ${\color{red}0.427} & -0.559$ \pm $0.172\\
kin8nm & 1.114$ \pm $0.0204 & {\color{red}1.283}$ \pm ${\color{red}0.0354} & $\mathbf{1.295}$$ \pm $$\mathbf{0.0312}$ & 0.999$ \pm $0.0687 & 0.979$ \pm $0.0513 & 1.178$ \pm $0.0264 & 1.223$ \pm $0.023\\
naval & {\color{red}6.544}$ \pm ${\color{red}0.0724} & 6.097$ \pm $0.628 & 6.483$ \pm $0.366 & 6.157$ \pm $0.235 & 5.944$ \pm $1.8 & $\mathbf{7.28}$$ \pm $$\mathbf{0.0598}$ & 5.651$ \pm $0.0445\\
\hline
\hline
\textit{Avg. ranking} & 5.39$ \pm $0.05 & 3.91$ \pm $0.06 & 2.82$ \pm $0.06 & 5.61$ \pm $0.07 & 3.70$ \pm $0.12 & $\mathbf{2.76}$$ \pm $$\mathbf{0.08}$ & 3.81$ \pm $0.07 \\
\hline
\end{tabular}
\end{minipage}\hfill

\centering
\caption{{\normalsize RMSE results in UCI datasets}} 
\begin{minipage}[c]{\columnwidth}
\centering
\tiny
\begin{tabular}{c|lll|ll|ll}
\hline
\textit{Dataset} & BBB & AVB & $\text{AADM}$ & $\text{fBNN}_\text{BNN}$ & VIP & $\text{SIP}_\text{BNN}$ & $\text{SIP}_\text{NS}$\\
\hline
boston & 3.52 $\pm$ 0.804 & {\color{red}2.53} $\pm$ {\color{red}0.461} & 2.55 $\pm$ 0.481 & 5.66 $\pm$ 10.2 & 2.62 $\pm$ 0.697 & 2.88 $\pm$ 0.616 & $\mathbf{2.29}$ $\pm$ $\mathbf{0.452}$\\
ccpp & 4.15 $\pm$ 0.201 & 4.05 $\pm$ 0.206 & 4.04 $\pm$ 0.207 & $\mathbf{3.77}$ $\pm$ $\mathbf{0.205}$ & 3.88 $\pm$ 0.224 & 3.92 $\pm$ 0.214 & {\color{red}3.87} $\pm$ {\color{red}0.198}\\
concrete & 6.64 $\pm$ 0.562 & 4.95 $\pm$ 0.569 & 4.92 $\pm$ 0.584 & 4.78 $\pm$ 0.694 & {\color{red}4.53} $\pm$ {\color{red}0.548} & 4.9 $\pm$ 0.825 & $\mathbf{4.45}$ $\pm$ $\mathbf{0.56}$\\
ee & 2.02 $\pm$ 0.231 & 1.32 $\pm$ 0.133 & 1.77 $\pm$ 0.208 & $\mathbf{0.639}$ $\pm$ $\mathbf{0.116}$ & {\color{red}0.789} $\pm$ {\color{red}0.13} & 1.47 $\pm$ 0.316 & 0.836 $\pm$ 0.106\\
wine & 0.649 $\pm$ 0.0388 & 0.633 $\pm$ 0.0342 & 0.631 $\pm$ 0.0339 & 0.78 $\pm$ 0.0645 & $\mathbf{0.622}$ $\pm$ $\mathbf{0.0437}$ & 0.636 $\pm$ 0.0367 & {\color{red}0.625} $\pm$ {\color{red}0.0452}\\
yatch & 2.3 $\pm$ 0.98 & 1.1 $\pm$ 0.42 & 1.3 $\pm$ 0.64 & 0.82 $\pm$ 0.31 & 0.58 $\pm$ 0.22 & $\mathbf{0.36}$ $\pm$ $\mathbf{0.11}$ & {\color{red}0.54} $\pm$ {\color{red}0.27}\\
kin8nm & 0.079 $\pm$ 0.0022 & {\color{red}0.067} $\pm$ {\color{red}0.0024} & $\mathbf{0.067}$ $\pm$ $\mathbf{0.0023}$ & 0.074 $\pm$ 0.0022 & 0.079 $\pm$ 0.0025 & 0.075 $\pm$ 0.0021 & 0.07 $\pm$ 0.0021\\
naval & ({\color{red}3.02} $\pm$ {\color{red}4.73})$e^{-5}$  & (4.96 $\pm$ 0.21) $e^{-5}$ & (3.61 $\pm$ 0.17)$e^{-5}$ & (3.78 $\pm$ 2.17)$e^{-5}$ & (3.18 $\pm$ 1.06)$e^{-5}$ & ($\mathbf{1.73}$ $\pm$ $\mathbf{0.15}$) $e^{-5}$ & (6.37 $\pm$ 0.98) $e^{-5}$\\
\hline
\hline
\textit{Avg. ranking} & 6.15$ \pm $0.04 & 4.14$ \pm $0.08 & 3.97$ \pm $0.09 & 3.82$ \pm $0.10 & 3.29$ \pm $0.09 & 3.53$ \pm $0.08 & $\mathbf{3.10}$$ \pm $$\mathbf{0.09}$ \\
\hline
\end{tabular}
\end{minipage}\hfill

\centering
\caption{{\normalsize CRPS results in UCI datasets}} 

\begin{minipage}[c]{\columnwidth}
\centering
\tiny
\begin{tabular}{c|lll|ll|ll}
\hline
 \textit{Dataset} & BBB & AVB & $\text{AADM}$ & $\text{fBNN}_\text{BNN}$ & VIP & $\text{SIP}_\text{BNN}$ & $\text{SIP}_\text{NS}$\\
\hline
boston & 1.859$ \pm $0.363 & 1.475$ \pm $0.281 & 1.456$ \pm $0.298 & 1.861$ \pm $1.137 & $\mathbf{1.348}$$ \pm $$\mathbf{0.241}$ & 1.573$ \pm $0.278 & {\color{red}1.45}$ \pm ${\color{red}0.259}\\
ccpp & 2.641$ \pm $0.103 & 2.615$ \pm $0.102 & 2.608$ \pm $0.104 & $\mathbf{2.013}$$ \pm $$\mathbf{0.08}$ & 2.229$ \pm $0.081 & {\color{red}2.114}$ \pm ${\color{red}0.082} & 2.2$ \pm $0.072\\
concrete & 3.755$ \pm $0.294 & 2.811$ \pm $0.262 & 2.615$ \pm $0.214 & {\color{red}2.35}$ \pm ${\color{red}0.301} & $\mathbf{2.273}$$ \pm $$\mathbf{0.196}$ & 2.485$ \pm $0.238 & 2.683$ \pm $0.24\\
ee & 1.121$ \pm $0.139 & 0.703$ \pm $0.07 & 0.694$ \pm $0.082 & {\color{red}0.41}$ \pm ${\color{red}0.03} & $\mathbf{0.387}$$ \pm $$\mathbf{0.061}$ & 0.532$ \pm $0.078 & 0.591$ \pm $0.078\\
wine & 0.362$ \pm $0.021 & 0.356$ \pm $0.019 & 0.363$ \pm $0.019 & 0.479$ \pm $0.049 & $\mathbf{0.341}$$ \pm $$\mathbf{0.02}$ & {\color{red}0.353}$ \pm ${\color{red}0.017} & 0.366$ \pm $0.028\\
yatch & 1.026$ \pm $0.185 & 0.475$ \pm $0.105 & 0.519$ \pm $0.128 & 0.838$ \pm $0.056 & {\color{red}0.212}$ \pm ${\color{red}0.071} & $\mathbf{0.154}$$ \pm $$\mathbf{0.03}$ & 0.384$ \pm $0.094\\
kin8nm & 0.047$ \pm $0.001 & {\color{red}0.042}$ \pm ${\color{red}0.002} & $\mathbf{0.041}$$ \pm $$\mathbf{0.001}$ & 0.042$ \pm $0.002 & 0.044$ \pm $0.001 & 0.042$ \pm $0.001 & 0.047$ \pm $0.002\\
naval & ({\color{red}1.6}$ \pm ${\color{red}0.28}) $e^{-4}$ & (3.46 $ \pm $1.56) $e^{-4}$ & (2.21$ \pm $ 1.22)$e^{-4}$ & (2.56$ \pm $ 0.95) $e^{-5}$ & (1.81$ \pm $ 0.68) $e^{-5}$ & ($\mathbf{9.4} \pm \mathbf{0.70}$) $e^{-5}$ & (6.8$ \pm $ 0.28) $e^{-5}$\\
\hline
\hline
\textit{Avg. ranking} & 5.84$ \pm $0.05 & 4.47$ \pm $0.08 & 4.03$ \pm $0.08 & 3.97$ \pm $0.07 & 2.57$ \pm $0.08 & $\mathbf{2.45}$$ \pm $$\mathbf{0.08}$ & 4.67$ \pm $0.08  \\
\hline
\end{tabular}
\end{minipage}\hfill

\end{table}

\normalsize

\end{document}